\documentclass{article} % For LaTeX2e
\usepackage{iclr2027_conference,times}

\usepackage{amsmath,amsfonts,bm}

\def\eqref#1{equation~\ref{#1}}
\def\1{\bm{1}}

\DeclareMathAlphabet{\mathsfit}{\encodingdefault}{\sfdefault}{m}{sl}
\SetMathAlphabet{\mathsfit}{bold}{\encodingdefault}{\sfdefault}{bx}{n}

\usepackage{hyperref}
\usepackage{url}
\usepackage{multirow}
\usepackage{booktabs}
\usepackage{graphicx}
\usepackage{subcaption}
\usepackage{amsmath}
\usepackage{ulem}
\usepackage{verbatim}
\usepackage{algorithm}
\usepackage{algpseudocode}
\usepackage{color}
\usepackage{ mathrsfs }

\newcommand{\mycomment}[1]{}

\usepackage{wrapfig}
\usepackage{algorithm}
\usepackage{algpseudocode}

\usepackage[table]{xcolor}
\usepackage{bm}
\usepackage{xfp}

\definecolor{heatGreen}{HTML}{B7DFB9}
\definecolor{heatRed}{HTML}{F2B0AA}

\newcommand{\heatrow}[2]{%
  \noalign{\gdef\heatmin{#1}\gdef\heatmax{#2}}%
}

\definecolor{hl}{RGB}{46,139,87}
\newcommand{\heatcap}{60}  % max green intensity (same as the PSNR/SSIM table)

\newcommand{\heatcell}[2]{%
  \edef\heatcommand{%
    \noexpand\cellcolor{hl!%
      \fpeval{round(\heatcap*(1-max(0,min(1,
        (ln(#1)+(#2)*ln(10)-ln(\heatmin))/
        (ln(\heatmax)-ln(\heatmin))
      ))),1)}%
    !white}%
  }%
  \heatcommand
}
\newcommand{\hval}[4][\relax]{%
  \heatcell{#2}{#4}%
  \if\relax\detokenize{#3}\relax
    $#1{#2}$%
  \else
    $#1{#2{\scriptstyle (#3){#4}}}$%
  \fi
}

\title{Source Anchoring for Physical Consistency in Flow Matching Models}

\author{%
Giulia Romoli\thanks{Equal contribution.}, \quad Filippo Ruffini\footnotemark[1] \\
Department of Diagnostics and Intervention, Ume\aa{} University, Sweden \\
\texttt{\{giulia.romoli, filippo.ruffini\}@umu.se}
\AND
Paolo Soda \\
Department of Diagnostics and Intervention, Ume\aa{} University, Sweden \\
Unit of Artificial Intelligence and Computer Systems, \\
Universit\`a Campus Bio-Medico di Roma, Italy \\
\texttt{p.soda@unicampus.it}
}

\iclrfinalcopy % Uncomment for camera-ready version, but NOT for submission.
\PassOptionsToPackage{table}{xcolor}  
\begin{document}

\maketitle
\fancyhead[L]{Submitted to ICLR 2027}
\begin{abstract}

Deep generative models are used to solve partial differential equations and model distributions of physical system states, but ensuring that the generated samples satisfy the governing laws remains challenging. 
Projection-based flow-matching methods enforce physics by correcting the flow from an unconstrained noise distribution.
These corrections shift the generated samples away from the distribution of target solutions, especially in high noise regions. 
To address this limitation, we propose Source Anchoring for Physical Consistency (SAPC), a Functional Flow Matching method that encodes the physical constraints into the source noise before generation begins.
We evaluate SAPC on five systems governed by partial differential equations, covering six tasks with linear and non-linear dynamics, and compare results against five baselines and the unconstrained backbone. 
Anchoring the source reduces the need for large corrections that drive samples onto admissible but off-distribution states, and SAPC reproduces the target distributions most accurately on every evaluated task, while matching the constraint precision of the best projection-based baselines. 
Ablation experiments show that this gain arises from pairing source projection with a matched training objective that regresses toward the projected source.
These results identify the source distribution as a key design choice for physically consistent generative modelling. 
\end{abstract}

\section{Introduction}

Partial Differential Equations (PDEs) govern the evolution of many physical phenomena, and solving them is crucial for numerous real-world applications~\citep{wang2023a, ji2025}.
Standard numerical methods approximate PDE solutions by discretising the continuous problem onto a spatio-temporal grid, at a cost that increases with finer resolutions or longer simulations, becoming prohibitive in many regimes~\citep{dai2026b}.
Machine learning has emerged as an alternative to conventional PDE solvers, with approaches ranging from Physics-Informed Neural Networks (PINNs) to Neural Operators (NOs)~\citep{li2021,lu2021}. 
These methods learn to derive a single solution for a given set of initial conditions (ICs), boundary conditions (BCs), and PDE parameters. 
In many settings, however, the problem setup is only partially specified, leaving unresolved degrees of freedom and multiple solutions compatible with the available information. 
Capturing this variability requires representing a range of possible solutions, which a single prediction cannot provide~\citep{dai2026a}.
This limitation motivates the use of generative models, which learn a conditional distribution over PDE solutions rather than a single input–output mapping. 
Sampling from this distribution yields multiple solutions that are consistent with the available information, and captures the variability arising from the incomplete knowledge of the system~\citep{wang2026}.
Among generative approaches, diffusion and flow-based models including Denoising Diffusion Probabilistic Models (DDPMs)~\citep{ho2020} and Flow Matching (FM)~\citep{lipman2023}, have become the primary framework for diverse applications~\citep{rombach2022,ho2022,kong2020,chamberlain2021}, given their ability to represent complex and high-dimensional distributions. 
Applied to PDEs, these models learn to transform noise into candidate solutions by training on a finite set of reference trajectories.
However, learning the statistical structure of the training samples does not guarantee that the generated outputs satisfy the governing PDE or its associated physical constraints~\citep{krishnapriyan2021,hansen2023}. 
Ensuring physical consistency therefore remains a key challenge when using diffusion and flow-matching models as samplers of PDE solutions.
\begin{figure}[t]
    \centering
    \includegraphics[width=0.8\textwidth]{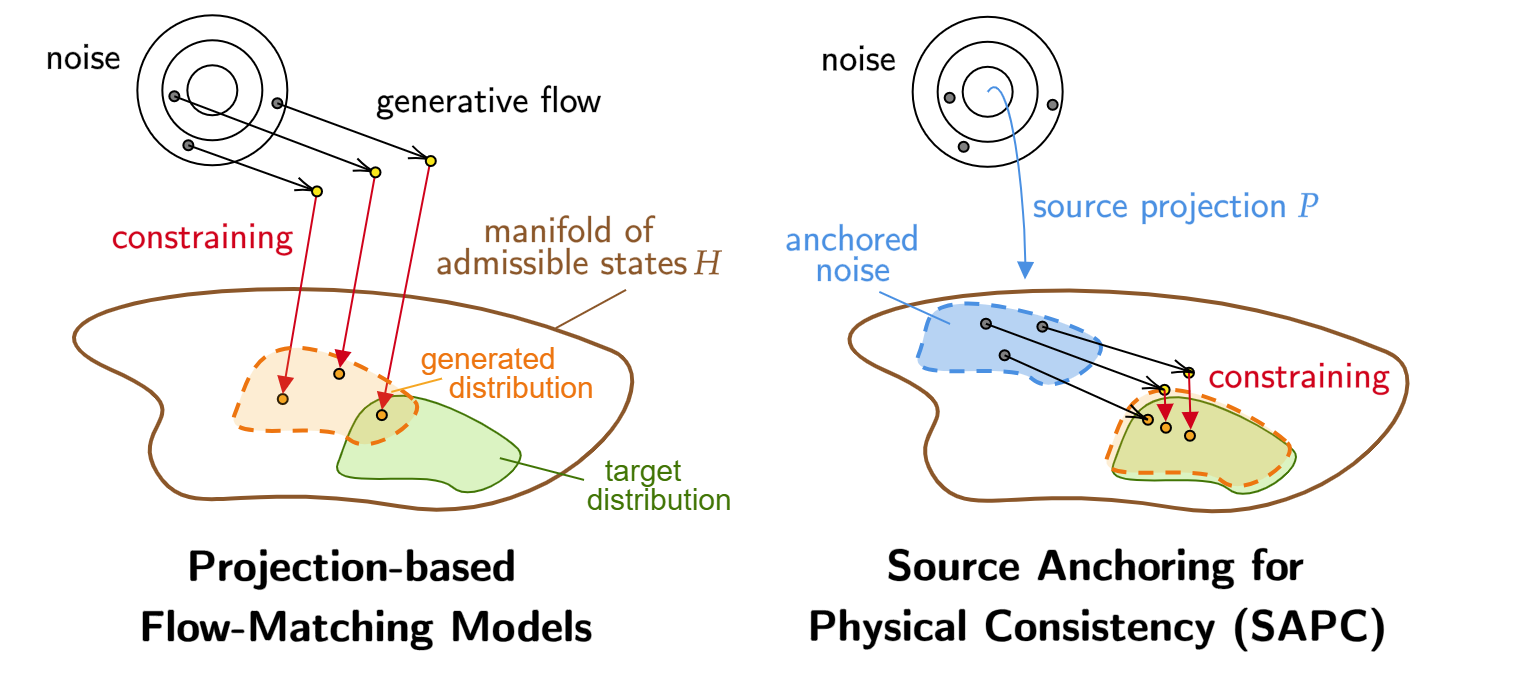}
    \caption{\textbf{Left:} existing projection-based flow-matching models correct the diffusion flow onto the manifold of admissible states to enforce physics, and the generated distribution (orange) may drift from the target one (green). \textbf{Right:} SAPC projects the noise on the manifold, anchoring it before diffusion starts (blue), mitigating the need for large corrections and reducing the distributional drift.}
    \label{fig:SAPC}
\end{figure}

Different strategies have been proposed to physically constrain diffusion, through training penalties~\citep{bastek2025,baldan2026}, architectural design~\citep{beucler2021}, gradient-based guidance~\citep{chung2023,huang2024,ben2024} and projection~\citep{christopher2024,utkarsh2025,cheng2025, christopher2026}.
Within this landscape, projection-based methods have emerged as a promising approach to enforce physical constraints, by mapping intermediate diffusion or final output samples onto the manifold of admissible solutions. 
These approaches differ in whether projection is incorporated during training or sampling and in where it acts along the generative trajectory. 
Despite these differences, current projection-based approaches share two main limitations.
First, projectors are constructed from the ICs, BCs, and integral conservation laws of the problem, without enforcing the dynamics encoded in the PDE~\citep{hansen2023,cheng2025,utkarsh2025,baldan2026}.
The resulting manifold therefore contains all the states that satisfy these constraints, including the true PDE solutions as well as states that are incompatible with the governing dynamics~\citep{rochman2025}.
Second, generation starts from random noise that does not obey the physical constraints and lies far from the manifold of admissible solutions~\citep{kynkaanniemi2024, rojas2026}.
When generation starts from an off-manifold source, the learned flow must recover the structure of the target distribution while also overcoming the violation of physical constraints and dynamics. 
Projection can mitigate these violations, but large corrections may move the samples away from the target distribution while enforcing the prescribed constraints, thus leading to solutions that are physically admissible but off-distribution with respect to the target ones (Figure~\ref{fig:SAPC}, left).

\begin{figure}[!h]
    \centering
    \includegraphics[width=\textwidth]{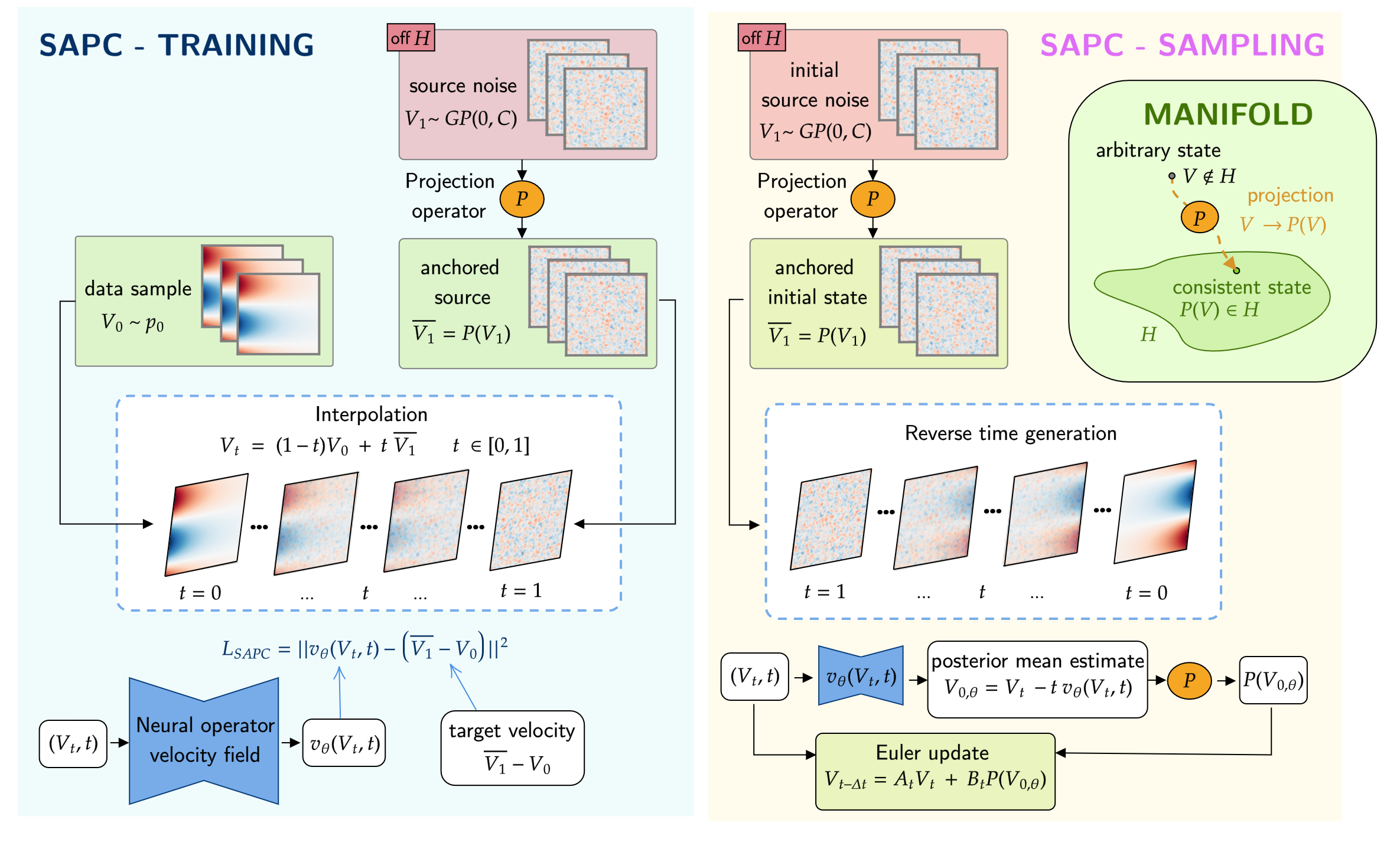}
    \caption{SAPC training (left) and sampling (right). SAPC anchors the source onto the physically-consistent manifold. The posterior-mean estimate is additionally projected at every Euler step. }
    \label{fig:method}
\end{figure}

In this work we propose Source Anchoring for Physical Consistency (SAPC), a Functional Flow Matching (FFM)~\citep{kerrigan2023} method that projects source samples onto the manifold of admissible solutions $\mathcal{H}$. This keeps the marginals close to $\mathcal{H}$ throughout the flow, limiting the need for large corrective projections and the associated distributional bias (Figure~\ref{fig:SAPC}, right).
The contributions of the work are the following: 

\begin{itemize}
\item We propose SAPC (Figure~\ref{fig:method}), a FFM method that anchors source samples onto the constraint manifold $\mathcal{H}$. 
During training, the network learns to transport anchored source samples to their paired target solutions; at sampling the estimated target solution is additionally projected at every integration step.

\item SAPC projects the source at both training and sampling time, and the endpoint estimate at every reverse-time integration step.
We show that source anchoring, paired with a training loss on the projected source, is what reduces the distributional bias, while endpoint projection drives constraint violations toward zero.

\item We evaluate SAPC on linear and non-linear PDE systems, yielding six tasks: Heat, Reaction–Diffusion, Stokes with initial or boundary conditions, Burgers, and Navier–Stokes. Against five baselines and unconstrained FFM, SAPC reduces distributional bias on every task, while matching the constraint precision of the best projection-based methods.
\end{itemize}

\section{Related work}
\label{sec:related}

Different strategies have been proposed to enforce physical knowledge into generative models.
Soft constraints augment the training loss with a penalty term proportional to the residual of the governing equation~\citep{li2024, bastek2025, li2025, baldan2026}, whereas hard constraints enforce the physical laws through dedicated output layers~\citep{beucler2021} or boundary-aware architectures~\citep{wang2024}.
Such approaches are tied to the selected system and must be redesigned for each new problem.
To improve generalization, a third family of methods enforces the constraint at sampling time, guiding a pre-trained model toward the physically consistent set of solutions.
Within this last family, two main categories have emerged.
Gradient-based methods steer the sampling trajectory by back-propagating the constraint residual, either by nudging each iterate in the direction that reduces the violation, like classifier guidance~\citep{chung2023, huang2024, yu2023}, or by optimizing the source point~\citep{ben2024}.
Projection-based methods instead introduce a projection step of the network's intermediate or denoised estimate onto the manifold of physically admissible states~\citep{hansen2023,cheng2025,utkarsh2025,baldan2026}.
Projection does not require calibrating a guidance weight and it avoids back-propagation, reducing computational cost.
Existing projection-based methods can be organized along three axes, based on \textit{where} the projection is applied along the diffusion trajectory, whether a \textit{training} phase is included, and \textit{how} the projection is computed with respect to the geometry of the manifold of physically-consistent states.
Along the first axis, the projection can act on the posterior-mean estimate~\citep{wang2023, cheng2025} or on the entire backward diffusion~\citep{christopher2024}.
The latter can distort the sampling trajectory when applied at high noise levels~\citep{kynkaanniemi2024, rojas2026}.
Annealing the projection strength across the diffusion trajectory has been proposed to mitigate this~\citep{narasimhan2025,genuist2026}.
The projection can also be folded into the training loss~\citep{li2026, christopher2026, baldan2026}.
Whether the projection acts at training or sampling time balances a trade-off between flexibility and consistency.
On the one hand, sampling-time projection operates zero-shot on a frozen backbone, and generalises to constraints not seen at training time, but the objective remains that of the unconstrained model and this can degrade sample quality~\citep{christopher2026}. 
On the other hand, training-time embedding ties the learned weights to the constraint seen during training, reducing generalizability.
Finally, when the manifold is a linear subspace, the projector is orthogonal and well-defined; when it is non-linear, the projector is locally approximated onto the tangent space, as done by PCFM~\citep{utkarsh2025}, which reports improved robustness over ECI~\citep{cheng2025} on problems with shocks or discontinuities.
Despite differences, all existing projection-based methods act on the diffusion trajectory, leaving source samples unconstrained. 
SAPC anchors the source onto the manifold and folds this into the training loss.
This design extends the first axis by introducing a method that projects the source, a setting left unexplored by prior work.
Being source projection folded into the training loss, the learned velocity field is aligned with the constrained target distribution.
On the third axis, SAPC is independent of the geometry of the constraint manifold and applies to linear and non-linear systems.

\section{Method}
\label{sec:method}

\textbf{Preliminaries.}
Many phenomena are governed by PDE systems of the generic form
$\mathcal{F}(\mathcal{V}) = 0$, where $\mathcal{V}:\Omega\to\mathbb{R}^{d}$ is a state of the system defined on the spatio-temporal
domain $\Omega$, and $\mathcal{F}$ is a differential operator encoding the physical law.
A state $\mathcal{V}$ encodes the spatial configuration of the system and its evolution over time, i.e., the full dynamics, and it is called a \textit{solution} of the PDE when it satisfies the selected law exactly, so that $\mathcal{F}(\mathcal{V}) = 0$.
The degree to which a state satisfies $\mathcal{F}(\mathcal{V}) = 0$ defines its \textit{physical consistency}.
This condition reduces to a finite set of algebraic checks on $\mathcal{V}$: compliance with the ICs, the BCs, and the integral conservation law $\int_{\Omega'} \mathcal{F}(\mathcal{V}) \, dx = 0$ obtained by integrating $\mathcal{F}(\mathcal{V}) = 0$ on subdomains $\Omega' \subseteq \Omega$.
To quantify deviations from ICs and BCs, we define a \textit{local} constraint $\mathcal{R}_{\text{loc}}(\mathcal{V})=\mathcal{V}|_{\Omega_{\text{loc}}}-g$, that measures how much $\mathcal{V}$ deviates from the prescribed IC/BC values $g$ on the subset $\Omega_{\text{loc}} \subset \Omega$ where they apply.
To check compliance with the integral conservation law, we also define a \textit{global} residual $\mathcal{R}_{\text{glob}}(\mathcal{V})=\int_{\Omega} \mathcal{F}(\mathcal{V}) \, dx - c$, that quantifies how much the integral departs from $c$, with $c$ the value the integral takes on any admissible solution.
We stack these into a constraint residual defined as $\mathcal{R}(\mathcal{V}) = [\mathcal{R}_{\text{loc}}(\mathcal{V}),\, \mathcal{R}_{\text{glob}}(\mathcal{V})]$.
From $\mathcal{R}(\mathcal{V})$, the manifold $\mathcal{H}$ of physically-consistent states is then defined as:
\begin{equation}
\mathcal{H} \;=\; \bigl\{\, \mathcal{V} : \Omega \to \mathbb{R}^{d} \;\bigm|\; \mathcal{R}(\mathcal{V}) = 0 \,\bigr\}
\label{eq:manifold}
\end{equation}

\noindent \textbf{Functional Flow Matching and constraint projection.}
We build our method upon the FFM framework~\citep{kerrigan2023} (detailed in Appendix~\ref{app:ffm}).
FFM learns a velocity field $v_{\theta}(\mathcal{V}_{t},t)$, where $\theta$ denotes the network’s trainable parameters, that transports the source noise distribution $p_{1}= \mathcal{GP}(0, C)$ at time $t=1$, defined as a Gaussian process on $L^{2}(\Omega)$ with zero mean and trace-class covariance operator $C$, to the target distribution of PDE solutions $p_{0}$ at $t=0$. 
During training, source $\mathcal{V}_{1}\sim p_{1}$ and target $\mathcal{V}_{0} \sim p_{0}$ samples are connected by the linear interpolant:
\begin{equation}
\mathcal{V}_{t} (\mathcal{V}_{0}, \mathcal{V}_{1}) \;=\; (1 - t)\,\mathcal{V}_{0} + t\,\mathcal{V}_{1},
\qquad v_{\theta}(\mathcal{V}_{t} \mid \mathcal{V}_{0}, \mathcal{V}_{1}) \;=\; \mathcal{V}_{1} - \mathcal{V}_{0}
\label{eq:ffm-interp}
\end{equation}
Standard FFM does not guarantee that generated samples lie on the constraint manifold $\mathcal{H}$, so we enforce $\mathcal{R}(\mathcal{V})=0$ through a projection operation.
Let $P : L^{2}(\Omega) \to \mathcal{H}$ denote the metric projection onto $\mathcal{H}$, defined as the optimization problem $P(\mathcal{V}) = \arg\min_{\mathcal{U}\in\mathcal{H}} \Vert{}\mathcal{U} - \mathcal{V}\Vert{}$.
The exact form of $P$ depends on the constraint function $\mathcal{R}(\mathcal{V})$ that defines $\mathcal{H}$. 
When $\mathcal{R}$ is affine, taking the form $\mathcal{R}(\mathcal{V}) = A\mathcal{V} - b$, being $A$ the linear operator encoding the constraints and $b$ the vector containing their prescribed values, then $P$ is orthogonal to the linear manifold and admits the closed-form $P(\mathcal{V}) = \mathcal{V} - A^{\top}(AA^{\top})^{-1}(A\mathcal{V} - b)$. 
For non-linear constraints $\mathcal{R}$, a closed-form projection is mathematically intractable. 
In these cases, $P$ is locally defined onto the tangent space of the curved manifold and approximated numerically using a damped Gauss-Newton algorithm~\citep{utkarsh2025}. 
The damping parameter ensures convergence, and the updates end once the predefined iteration budget has been reached.
Non-linear projection settings are specified in Appendix, Section~\ref{app:projector}.

\noindent \textbf{Source Anchoring for Physical Consistency.}
As shown in Figure~\ref{fig:method}, SAPC enforces physical consistency on the FFM backbone by projecting two objects onto $\mathcal{H}$: the source draw $\mathcal{V}_{1}$ at both training and sampling time, and the posterior-mean estimate $\mathcal{V}_{0,\theta}$ at every reverse-time step.
During training, the source draw is replaced with its projection onto $\mathcal{H}$, $\bar{\mathcal{V}}_{1} = P(\mathcal{V}_{1})$,
and the FFM is trained to regress the velocity between
$\mathcal{V}_{0}$ and $\bar{\mathcal{V}}_{1} $:
\begin{equation}
\mathcal{L}_{\text{SAPC}} \;=\; \mathbb{E}_{t,\,\mathcal{V}_{0},\,\mathcal{V}_{1}}\,
\bigl\| v_{\theta}(\mathcal{V}_{t}, t)
      - \bigl(\bar{\mathcal{V}}_{1} - \mathcal{V}_{0}\bigr) \bigr\|^{2},
\qquad
\mathcal{V}_{t} \;=\; (1 - t)\,\mathcal{V}_{0} + t\, \bar{\mathcal{V}}_{1}
\label{eq:sapc-loss}
\end{equation}
Training is summarised in Algorithm~\ref{alg:sapc-training}.
Both endpoints of the diffusion flow lie on $\mathcal{H}$, the target $\mathcal{V}_{0}$ by construction, the source $\bar{\mathcal{V}}_{1}$ by projection. 
The residual along the flow depends on the geometry of $\mathcal{H}$. 
On the one hand, when $\mathcal{R}$ is affine, it follows from linearity that $\mathcal{R}(\mathcal{V}_{t})= (1-t)\,\mathcal{R}(\mathcal{V}_{0}) + t\,\mathcal{R}(\bar{\mathcal{V}}_{1})= 0$ for every $t \in [0,1]$ and the entire diffusion trajectory lies on $\mathcal{H}$. 
On the other hand, when $\mathcal{R}$ is non-linear, the residual no longer vanishes along the flow but remains bounded:
\begin{equation}
\bigl\|\mathcal{R}(\mathcal{V}_{t})\bigr\|
\;\leq\; \tfrac{1}{2}\, t(1-t)\, L\,
\bigl\|\mathcal{V}_{0} - \bar{\mathcal{V}}_{1}\bigr\|^{2},
\qquad
L \;=\; \sup_{\mathcal{V}\in[\mathcal{V}_{0},\bar{\mathcal{V}}_{1}]}
\bigl\|\nabla^{2}\mathcal{R}(\mathcal{V})\bigr\|
\label{eq:sapc-nonlinear-interp}
\end{equation}
The derivation of this last equation is addressed in Appendix~\ref{app:residual-bounds}.
From equation~\ref{eq:sapc-nonlinear-interp}, the residual vanishes at both endpoints and grows at most quadratically in between, with a factor $t(1-t)$ that peaks at $t = 1/2$ and decays symmetrically to zero at both ends.
In both cases, anchoring the source onto $\mathcal{H}$ keeps the flow near it.
As derived in Appendix~\ref{app:residual-bounds}, the interpolant marginals, i.e., the distributions of $\mathcal{V}_{t}$ as $t$ varies over $[0,1]$, stay close to $\mathcal{H}$ throughout.
Without source anchoring, by contrast, $\mathcal{R}(\mathcal{V}_{1}) \neq 0$. 
The residual along the interpolant is $\|\mathcal{R}(\mathcal{V}_{t})\| = t\,\|\mathcal{R}(\mathcal{V}_{1})\|$ when $\mathcal{H}$ is a linear subspace, and $\|\mathcal{R}(\mathcal{V}_{t})\| \leq t\,\|\mathcal{R}(\mathcal{V}_{1})\| + \mathcal{O}\bigl(t(1-t)\bigr)$ when it is curved.
In both cases, $\|\mathcal{R}(\mathcal{V}_{1})\| \neq 0$ at $t=1$, so the network has to transport a source that sits a fixed distance off $\mathcal{H}$ onto it during training, rather than refining one that already lies on (or is close to) it. 

During sampling, $\mathcal{V}_{1}$ is again substituted by $\bar{\mathcal{V}}_{1} = P(\mathcal{V}_{1})$, but projecting the source pins only one endpoint of the flow.
The reverse-time iterates $\mathcal{V}_{t}$ can still drift off $\mathcal{H}$, and the posterior-mean estimate $\mathcal{V}_{0,\theta} = \mathcal{V}_{t} - t\, v_\theta(\mathcal{V}_{t}, t)$ inherits any residual bias of the learned velocity field $v_\theta$. 
During sampling, we therefore replace $\mathcal{V}_{0,\theta}$ with its projection $\bar{\mathcal{V}}_{0,\theta}=P(\mathcal{V}_{0,\theta})$ at each generative step, to steer the update along a constraint-satisfying direction:
\begin{equation}
\mathcal{V}_{t-\Delta t} = A_t\,\mathcal{V}_{t} + B_t\,\bar{\mathcal{V}}_{0,\theta},
\quad A_t = \frac{t-\Delta t}{t},
\quad B_t = \frac{\Delta t}{t}
\label{eq:sapc-sampling}
\end{equation}
The full sampling procedure is summarised in Algorithm~\ref{alg:sapc-sampling}.
Endpoint projection is complementary to source anchoring. 
The latter keeps the training interpolant on or near $\mathcal{H}$ throughout the flow, so the velocity field is regressed on near-consistent states; endpoint projection instead corrects the residual drift that remains along each Euler step. 
Those sampling-time corrections have been reported to work better when the iterate is already close to the target manifold~\citep{kynkaanniemi2024, rojas2026}, as distortions can be introduced when projection is applied at high noise levels.

\begin{figure}[!h]
\begin{minipage}[t]{0.58\textwidth}
\begin{algorithm}[H]
\caption{SAPC Training}
\label{alg:sapc-training}
\small
\begin{algorithmic}[1]
\Require Data $p_{\mathcal{V}}$, source $\mathcal{GP}(0,C)$,
    projection $P$, network $v_{\theta}$, steps $N$
\For{$n = 1$ \textbf{to} $N$}
    \State Sample $\mathcal{V}_{0} \sim p_{0}$,
           $\mathcal{V}_{1} \sim \mathcal{GP}(0,C)$,
           $t \sim \mathcal{U}[0,1]$
    \State $\bar{\mathcal{V}}_{1} \gets P(\mathcal{V}_{1})$
    \State $\mathcal{V}_{t} \gets
           (1-t)\mathcal{V}_{0}+t\bar{\mathcal{V}}_{1}$
    \State $\mathcal{L} \gets
           \|v_{\theta}(\mathcal{V}_{t},t)
             -(\bar{\mathcal{V}}_{1}-\mathcal{V}_{0})\|^{2}$
    \State Update $\theta$ by gradient descent on $\mathcal{L}$
\EndFor
\end{algorithmic}
\end{algorithm}
\end{minipage}\hfill
\begin{minipage}[t]{0.38\textwidth}
\begin{algorithm}[H]
\caption{SAPC Sampling (Euler)}
\label{alg:sapc-sampling}
\small
\begin{algorithmic}[1]
\Require Velocity field $v_{\theta}$, projection $P$, steps $N$
\State Sample $\mathcal{V}_{1} \sim \mathcal{GP}(0, C)$
\State $\mathcal{V}_{1} \gets P(\mathcal{V}_{1})$
\For{$i = N, N-1, \dots, 1$}
    \State $\mathcal{V}_{0,\theta} \gets
           \mathcal{V}_{t} - t\,v_{\theta}(\mathcal{V}_{t}, t)$
    \State $\mathcal{V}_{t -\Delta t} \gets
           A_t\,\mathcal{V}_{t} + B_t\,P(\mathcal{V}_{0,\theta})$
\EndFor
\State \Return $\mathcal{V}_{0}$
\end{algorithmic}
\end{algorithm}
\end{minipage}
\end{figure}

\section{Empirical evaluation}
\label{sec:experiments}

%dataset table
\begin{table}[!h]
\centering
\caption{Datasets specifications: dimensionality; spatial and temporal domain; spatial and temporal resolution (number of discretization points); local constraint and global conservation law. For the local constraint, either IC/BC or a distributional parameter is held out (like initial vorticity for Navier-Stokes). 
For the global conservation law either LMC/NLMC, NLEC, or LVC.}
\label{tab:datasets}
\begin{tabular}{cccccc}
\toprule
\textbf{Dataset} & \textbf{Dimensionality} & \textbf{Domain} & \textbf{Resolution} & \textbf{Local} & \textbf{Global}\\
\midrule
Heat & 1D & $[0,2\pi]\times[0,1]$ & $100\times100$ & IC & LMC\\
\midrule
RD & 1D & $[0,1)\times[0,1]$ & $128\times100$ & IC & NLMC\\
\midrule
\multirow{2}{*}{Stokes}
    & \multirow{2}{*}{1D} & \multirow{2}{*}{$[0,1]\times[0,1]$} & \multirow{2}{*}{$100\times100$} & IC & \multirow{2}{*}{NLEC}\\
    & & & & BC & \\
\midrule
Burgers & 1D & $[0,1)\times[0,1]$ & $100\times100$ & IC & LMC\\
\midrule
NS & 2D & $[0,1]^2\times[0,49]$ & $[64,64]\times50$ & init. vorticity & LVC\\
\bottomrule
\end{tabular}
\end{table}

We evaluate SAPC on five PDE systems from the literature~\citep{cheng2025, utkarsh2025, christopher2026}, spanning linear and non-linear dynamics and summarised in Table~\ref{tab:datasets}: 1D Heat, 1D Reaction-Diffusion (RD), 1D Stokes, 1D Burgers, and 2D Navier-Stokes (NS).
Each PDE is paired with a local constraint, defined by an IC, a BC, or a distributional parameter; and a global constraint, involving Linear or Non-Linear Mass Conservation (LMC, NLMC), Non-Linear Energy Conservation (NLEC), or Linear Vorticity Conservation (LVC).
The governing equations, the definition of the manifold $\mathcal{H}$ that derives from the chosen constraints, and the sampled parameter ranges are given in Appendix~\ref{app:datasets}.
For each dataset we generate $6400$ training trajectories ($10^4$ for NS), $1225$ validation trajectories, and four disjoint test sets of $1225$ trajectories each. 
The test sets differ only in the local constraint (IC, BC, or distributional parameter) held fixed within the set.
To assess stochastic variability, every configuration is drawn under three seeds $\{0,\,42,\,205\}$, and all reported metrics are means $\pm$~standard errors across the $3\times 4$ combinations of seed and test sets.
We compare SAPC with six competitors that represent the current state of the art in physics-constrained modelling: unconstrained FFM~\citep{kerrigan2023} as a lower bound; gradient-based D-Flow~\citep{ben2024} (source optimization); soft-constraining PBFM~\citep{baldan2026} (residual penalty at training); and three sampling-time projection methods, ECI~\citep{cheng2025}, PCFM~\citep{utkarsh2025}, and CAFM~\citep{christopher2026}, with CAFM additionally folding the projection into the training loss. 
All methods share the same FFM backbone, a Fourier Neural Operator~\citep{li2021} regressing the conditional velocity $v_{\theta}$, and the same projector on each dataset.
The architecture and other optimization settings are further discussed in Appendix~\ref{app:setup}, whilst the implementation of each competitor is detailed in Appendix~\ref{app:competitors}.
For each method we report the pointwise mean and standard deviation for the Mean Square Error (MMSE, SMSE) between generated and reference trajectories. 
We measure distribution fidelity via the Fr\'echet Poseidon Distance (FPD), computed on the features extracted from the trajectories by the Poseidon-B encoder~\citep{herde2024}, and the Wasserstein distance $W_2$, which complements FPD by measuring distributional fidelity pointwise rather than through pooled features.
Constraint satisfaction via the constraint error is computed separately on the local and global residuals, $\mathrm{CE}_{L} = \tfrac{1}{N}\sum_{n=1}^{N} \|\mathcal{R}_{\text{loc}}(\mathcal{V}_{n})\|_{2}$ and $\mathrm{CE}_{G} = \tfrac{1}{N}\sum_{n=1}^{N} \|\mathcal{R}_{\text{glob}}(\mathcal{V}_{n})\|_{2}$, and summarized as $\mathrm{CE}$, computed as the mean of $\mathrm{CE}_{L}$ and $\mathrm{CE}_{G}$, normalized to the unconstrained FFM ($\mathrm{CE}=1$ for it). 
An extended definition of each metric is given in Appendix~\ref{app:metrics}. 

\begin{table}[ht]
\centering
\caption{Generation on PDEs with linear and non-linear dynamics.
Each cell reports the mean $m$ and standard error $s$ across seeds and held-out test sets in the compact form $m(s)p$, meaning $(m \pm s) \times 10^{p}$. 
For example, $2.2(0.8)-5$ denotes $(2.2 \pm 0.8) \times 10^{-5}$.
Lower values indicate better performance, with best result in bold and second best underlined.
Cell shading ranges from dark green (lowest mean) to white (highest mean) on a logarithmic scale, with normalization applied within each row.}
\label{tab:generative_performance}
\setlength{\tabcolsep}{1.4pt}

\begin{tabular}{rcccccccc}
\toprule
 & \small Metric & SAPC & PCFM & ECI & CAFM & PBFM & D-Flow & FFM \\
\midrule

\heatrow{2.2e-5}{8.3e-2}
\multirow{4}{*}{\rotatebox[origin=c]{90}{\small Heat}}
 & \small MMSE $\downarrow$
 & \hval[\bm]{2.2}{0.8}{-05}
 & \hval[\underline]{1.2}{0.2}{-02}
 & \hval{1.9}{0.6}{-02}
 & \hval{2.2}{0.6}{-02}
 & \hval{5.8}{1.3}{-02}
 & \hval{8.3}{6.0}{-02}
 & \hval{5.7}{1.3}{-02} \\
\heatrow{2.3e-5}{1.5e1}
 & \small SMSE $\downarrow$
 & \hval[\bm]{2.3}{0.4}{-05}
 & \hval{6.4}{0.5}{-02}
 & \hval{7.7}{3.0}{-02}
 & \hval{1.2}{0.3}{-01}
 & \hval{3.8}{0.1}{-02}
 & \hval{1.5}{1.2}{+01}
 & \hval[\underline]{3.7}{0.1}{-02} \\
\heatrow{4.6e-3}{1.5e1}
 & \small FPD $\downarrow$
 & \hval[\bm]{4.6}{1.3}{-03}
 & \hval{5.7}{0.9}{+00}
 & \hval{7.7}{2.4}{+00}
 & \hval{1.1}{0.3}{+01}
 & \hval{2.8}{0.3}{+00}
 & \hval{1.5}{0.2}{+01}
 & \hval[\underline]{2.7}{0.3}{+00} \\
\heatrow{2.0e-6}{2.2}
 & \small  $\mathrm{CE}$ $\downarrow$
 & \hval[\bm]{2.0}{0.0}{-06}
 & \hval[\underline]{2.0}{0.0}{-06}
 & \hval{2.6}{0.1}{-06}
 & \hval{2.2}{0.1}{-06}
 & \hval{6.5}{0.3}{-01}
 & \hval{2.2}{0.9}{+00}
 & \hval{1}{}{0} \\
\midrule

\heatrow{5.8e-5}{5.6e-2}
\multirow{4}{*}{\rotatebox[origin=c]{90}{RD}}
 & \small MMSE $\downarrow$
 & \hval[\bm]{5.8}{0.9}{-05}
 & \hval{1.9}{0.6}{-02}
 & \hval[\underline]{5.8}{1.6}{-03}
 & \hval{2.3}{0.7}{-02}
 & \hval{4.1}{1.1}{-02}
 & \hval{5.6}{4.0}{-02}
 & \hval{4.2}{1.2}{-02} \\
\heatrow{1.9e-5}{4.8e1}
 & \small SMSE $\downarrow$
 & \hval[\bm]{1.9}{0.2}{-05}
 & \hval{9.4}{1.5}{-02}
 & \hval{3.4}{0.2}{-02}
 & \hval{5.5}{0.1}{-02}
 & \hval{3.4}{0.0}{-02}
 & \hval{4.8}{4.4}{+01}
 & \hval[\underline]{3.2}{0.1}{-02} \\
\heatrow{5.9e-1}{1.8e2}
 & \small FPD $\downarrow$
 & \hval[\bm]{5.9}{2.0}{-01}
 & \hval{1.8}{0.3}{+02}
 & \hval{1.4}{0.2}{+02}
 & \hval{1.6}{0.2}{+02}
 & \hval{1.2}{0.3}{+02}
 & \hval[\underline]{4.1}{0.6}{+01}
 & \hval{1.2}{0.3}{+02} \\
\heatrow{3.3e-6}{3.1e1}
 & \small  $\mathrm{CE}$ $\downarrow$
 & \hval[\bm]{3.3}{0.2}{-06}
 & \hval{3.3}{0.2}{-06}
 & \hval{3.6}{0.2}{-06}
 & \hval[\underline]{3.3}{0.2}{-06}
 & \hval{6.9}{0.2}{-01}
 & \hval{3.1}{2.7}{+01}
 & \hval{1}{}{0} \\
\midrule

\heatrow{3.0e-3}{1.5e-2}
\multirow{4}{*}{\rotatebox[origin=c]{90}{\small Stokes IC}}
 & \small MMSE $\downarrow$
 & \hval[\bm]{3.0}{1.2}{-03}
 & \hval{1.1}{0.1}{-02}
 & \hval{9.8}{2.5}{-03}
 & \hval{1.5}{0.2}{-02}
 & \hval{1.4}{0.4}{-02}
 & \hval[\underline]{6.3}{3.2}{-03}
 & \hval{1.5}{0.4}{-02} \\
\heatrow{1.1e-3}{8.7e-1}
 & \small SMSE $\downarrow$
 & \hval[\bm]{1.1}{0.4}{-03}
 & \hval{2.7}{0.3}{-02}
 & \hval[\underline]{1.2}{0.2}{-02}
 & \hval{2.7}{0.3}{-02}
 & \hval{1.9}{0.2}{-02}
 & \hval{8.7}{7.4}{-01}
 & \hval{1.9}{0.2}{-02} \\
\heatrow{2.8e-1}{7.9}
 & \small FPD $\downarrow$
 & \hval[\bm]{2.8}{1.2}{-01}
 & \hval{3.6}{0.4}{+00}
 & \hval[\underline]{2.0}{0.2}{+00}
 & \hval{3.7}{0.5}{+00}
 & \hval{7.8}{2.1}{+00}
 & \hval{3.8}{1.6}{+00}
 & \hval{7.9}{2.5}{+00} \\
\heatrow{5.0e-2}{1.0e1}
 & \small  $\mathrm{CE}$ $\downarrow$
 & \hval[\underline]{6.1}{3.3}{-02}
 & \hval{6.2}{0.9}{-01}
 & \hval{1.8}{0.5}{-01}
 & \hval[\bm]{5.0}{1.6}{-02}
 & \hval{6.9}{0.5}{-01}
 & \hval{1.0}{0.8}{+01}
 & \hval{1}{}{0} \\
\midrule

\heatrow{1.4e-3}{1.1e-1}
\multirow{4}{*}{\rotatebox[origin=c]{90}{\small Stokes BC}}
 & \small MMSE $\downarrow$
 & \hval[\bm]{1.4}{0.1}{-03}
 & \hval{1.4}{0.3}{-02}
 & \hval[\underline]{7.4}{1.4}{-03}
 & \hval{4.4}{0.3}{-02}
 & \hval{5.3}{0.8}{-02}
 & \hval{1.1}{0.7}{-01}
 & \hval{5.3}{0.8}{-02} \\
\heatrow{3.8e-3}{1.9e1}
 & \small SMSE $\downarrow$
 & \hval[\bm]{3.8}{0.4}{-03}
 & \hval{7.5}{4.7}{-01}
 & \hval[\underline]{7.3}{0.8}{-03}
 & \hval{2.1}{0.1}{-02}
 & \hval{3.5}{0.1}{-02}
 & \hval{1.9}{1.6}{+01}
 & \hval{3.2}{0.0}{-02} \\
\heatrow{1.7}{1.8e1}
 & \small FPD $\downarrow$
 & \hval[\bm]{1.7}{0.2}{+00}
 & \hval{4.7}{1.5}{+00}
 & \hval{2.5}{0.9}{+00}
 & \hval{9.6}{1.0}{+00}
 & \hval{2.2}{0.4}{+00}
 & \hval{1.8}{0.3}{+01}
 & \hval[\underline]{2.1}{0.5}{+00} \\
\heatrow{1.1e-6}{2.4e2}
 & \small  $\mathrm{CE}$ $\downarrow$
 & \hval{3.6}{0.4}{-02}
 & \hval{4.6}{0.4}{-01}
 & \hval[\underline]{2.7}{0.4}{-02}
 & \hval[\bm]{1.1}{0.2}{-06}
 & \hval{6.7}{0.2}{-01}
 & \hval{2.4}{2.1}{+02}
 & \hval{1}{}{0} \\
\midrule

\heatrow{6.4e-5}{9.0e-2}
\multirow{4}{*}{\rotatebox[origin=c]{90}{\small Burgers}}
 & \small MMSE $\downarrow$
 & \hval[\bm]{6.4}{1.8}{-05}
 & \hval{3.0}{0.4}{-02}
 & \hval{1.5}{0.2}{-02}
 & \hval{2.0}{0.2}{-02}
 & \hval{8.8}{1.3}{-02}
 & \hval[\underline]{1.5}{0.2}{-03}
 & \hval{9.0}{1.2}{-02} \\
\heatrow{5.5e-5}{1.8e-1}
 & \small SMSE $\downarrow$
 & \hval[\bm]{5.5}{1.1}{-05}
 & \hval{1.6}{0.3}{-01}
 & \hval{1.8}{0.6}{-01}
 & \hval{7.6}{0.3}{-02}
 & \hval{4.9}{0.1}{-02}
 & \hval[\underline]{1.6}{0.5}{-02}
 & \hval{5.0}{0.1}{-02} \\
\heatrow{1.5e-2}{1.2e1}
 & \small FPD $\downarrow$
 & \hval[\bm]{1.5}{0.4}{-02}
 & \hval{1.1}{0.3}{+01}
 & \hval{1.2}{0.5}{+01}
 & \hval{3.4}{0.2}{+00}
 & \hval{2.8}{0.4}{+00}
 & \hval[\underline]{1.7}{0.3}{+00}
 & \hval{2.8}{0.4}{+00} \\
\heatrow{1.4e-6}{1.0}
 & \small  $\mathrm{CE}$ $\downarrow$
 & \hval{1.7}{0.2}{-06}
 & \hval[\underline]{1.6}{0.0}{-06}
 & \hval{2.0}{0.2}{-06}
 & \hval[\bm]{1.4}{0.0}{-06}
 & \hval{6.4}{0.1}{-01}
 & \hval{3.4}{0.4}{-01}
 & \hval{1}{}{0} \\
\midrule

\heatrow{5.3e-2}{4.4e-1}
\multirow{4}{*}{\rotatebox[origin=c]{90}{\small NS}}
 & \small MMSE $\downarrow$
 & \hval[\bm]{5.3}{0.6}{-02}
 & \hval{2.2}{0.0}{-01}
 & \hval{4.4}{0.1}{-01}
 & \hval{2.2}{0.1}{-01}
 & \hval[\underline]{1.8}{0.1}{-01}
 & \hval{2.3}{1.7}{-01}
 & \hval[\underline]{1.8}{0.1}{-01} \\
\heatrow{3.1e-2}{1.5e1}
 & \small SMSE $\downarrow$
 & \hval[\bm]{3.1}{0.4}{-02}
 & \hval{1.7}{0.1}{-01}
 & \hval{3.1}{0.1}{-01}
 & \hval{1.7}{0.1}{-01}
 & \hval[\underline]{6.3}{0.3}{-02}
 & \hval{1.5}{1.4}{+01}
 & \hval{6.9}{0.3}{-02} \\
\heatrow{1.1}{5.6}
 & \small FPD $\downarrow$
 & \hval[\bm]{1.1}{0.1}{+00}
 & \hval{3.0}{0.1}{+00}
 & \hval{3.2}{0.1}{+00}
 & \hval{2.8}{0.1}{+00}
 & \hval{2.9}{0.1}{+00}
 & \hval{5.6}{1.8}{+00}
 & \hval[\underline]{2.7}{0.1}{+00} \\
\heatrow{7.0e-7}{1.2}
 & \small  $\mathrm{CE}$ $\downarrow$
 & \hval{7.3}{0.1}{-07}
 & \hval[\bm]{7.0}{0.0}{-07}
 & \hval{7.2}{0.1}{-07}
 & \hval[\underline]{7.0}{0.1}{-07}
 & \hval{8.5}{0.4}{-01}
 & \hval{1.2}{0.3}{+00}
 & \hval{1}{}{0}  \\
\bottomrule
\end{tabular}
\end{table}

\begin{figure}[!h]
    \centering
    \includegraphics[width=\textwidth]{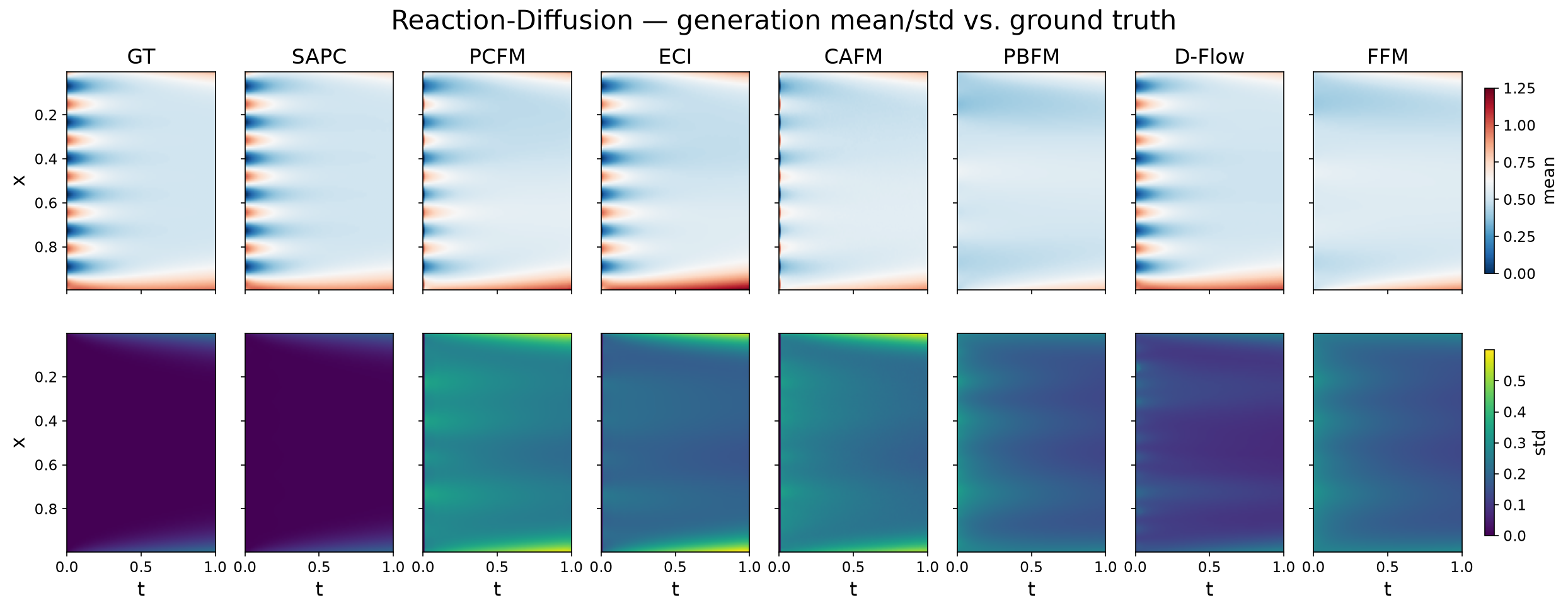}
    \caption{Pointwise mean (top) and standard deviation (bottom) over generated test RD trajectories. }
    \label{fig:baseline_comparison_rd}
\end{figure}

\noindent \textbf{Distributional fidelity.} Table~\ref{tab:generative_performance} shows that SAPC achieves the closest distributional match to the reference, both on the first two marginal moments (MMSE, SMSE) and on FPD, on all the datasets presented in this work.
These results suggest that anchoring the source distribution to $\mathcal{H}$ helps the learned flow recover the statistical structure of the reference distribution more faithfully.
By contrast, competing methods leave the source noise unconstrained, requiring the flow to bridge the mismatch between physically inconsistent source samples and admissible target solutions. 
Resolving this mismatch requires large corrective projections that distort the generated distribution, leading to lower distributional metrics.
Figure~\ref{fig:baseline_comparison_rd} shows that SAPC best reproduces the pointwise mean (top) and standard deviation (bottom) of the ground-truth trajectories (GT, first column) for the RD dataset (other datasets in Appendix~\ref{app:additional_results}).
SAPC improves distributional accuracy on every task, although the magnitude of the gain is dataset-dependent. 
For each distributional metric and relative to the second-best method, SAPC reduces errors by approximately one to three orders of magnitude on Heat, RD, and Burgers, with smaller gains on the Stokes tasks and 2D NS (Table~\ref{tab:generative_performance}).
This trend reflects how strongly the constraint $\mathcal{R}(\mathcal{V}) = 0$ restricts the range of possible solutions in each dataset.
For Heat, the IC and mass conservation leave only the diffusivity unspecified within the set of PDE solutions, making the choice of the constraint-enforcement mechanism a key factor in the final result.
For NS, fixing the initial vorticity and conserving total vorticity constrains only a limited part of the dynamics, leaving much of the two-dimensional turbulent evolution unspecified by these constraints. 
All methods must therefore learn the remaining dynamical structure from data, limiting the improvement that constraint enforcement alone can provide.
As the constraint no longer determines most of the solution, the choice of enforcement mechanism becomes less critical, and the difference between SAPC and competing methods narrows down.

\begin{figure}[!h]
    \centering
    \includegraphics[width=\textwidth]{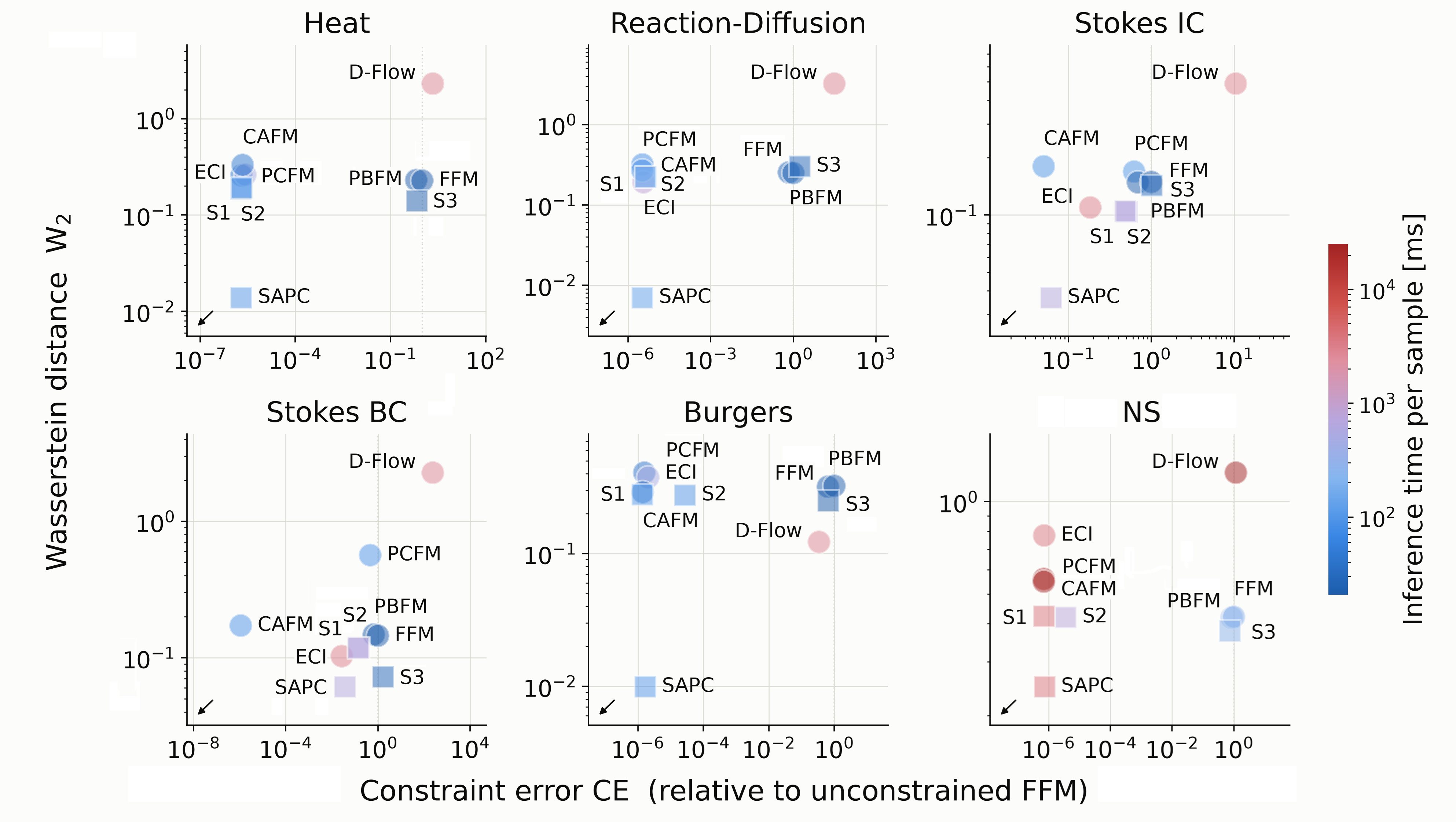}
\caption{Wasserstein distance $W_2$ against the constraint error $\mathrm{CE}$ (both on log scales), with marker colour encoding wall-clock inference time per sample. 
Arrows point to the bottom-left corner, where both metrics are jointly minimised.
Squares are SAPC and its ablated variants (S1, S2, and S3).}
    \label{fig:performance}
\end{figure}

\noindent \textbf{Constraint errors.} 
Figure~\ref{fig:performance} compares all methods across the datasets, showing the Wasserstein distance $W_2$ against the constraint error ($\mathrm{CE}$), normalised by the unconstrained FFM baseline. 
Both axes use logarithmic scales, and marker colours indicate inference time per sample. 
SAPC achieves the lowest $W_2$ across all datasets, extending the distributional improvements observed in MMSE, SMSE, and FPD (Table~\ref{tab:generative_performance}; additional results in Appendix~\ref{app:additional_results}, Table~\ref{tab:generative_performance_supp}).
Two patterns emerge among the competing methods. 
On the one hand, PCFM, ECI, and CAFM use the same sampling-time projector as SAPC and achieve comparable  $\mathrm{CE}$, on the order of $10^{-6}$–$10^{-7}$, on Heat, RD, Burgers, and NS. 
However, their $W_2$ values are one to two orders of magnitude higher on Heat, RD and Burgers, and still two to three times higher on NS, showing that comparable constraint satisfaction does not imply comparable agreement with the reference distribution.
On the other hand, PBFM and D-Flow remain broadly comparable to unconstrained FFM in both  $\mathrm{CE}$ and $W_2$, indicating that neither training-time penalties (PBFM) nor inference-time guidance (D-Flow) achieve the same level of constraint satisfaction and distributional fidelity as SAPC.
Stokes BC is the only 1D benchmark where the competing methods show less separation in distributional accuracy. Here, the local constraint specifies the boundary at $x=0$ rather than the initial spatial profile at $t=0$. Because the Stokes field decays exponentially with distance, fixing the boundary at $x=0$ constrains a substantial part of this variability, reducing the differences in distributional accuracy among methods.
Stokes BC is also the one task where SAPC does not match the best competitor on  $\mathrm{CE}$. 
Table~\ref{tab:generative_performance_supp} (Appendix~\ref{app:additional_results}) shows the gap is concentrated in the global constraint error.
Nonetheless, this residual lies at the floor of the discretised energy balance, i.e., the error measured on the exact solutions.
Finally, regarding performance, SAPC has an inference cost comparable to that of the other methods on the 1D benchmarks, as indicated by the marker colours in Figure~\ref{fig:performance}.
On NS, projection-based methods require comparable inference times, with most of the computational cost arising from projection over the three-dimensional space–time domain. A detailed analysis is provided in Appendix~\ref{app:performance}.

\noindent \textbf{Ablation study.}
Figure~\ref{fig:performance} also presents three ablated variants of SAPC (named S1, S2, and S3; square markers), designed to assess which part of SAPC contributes to the observed improvements.
S1 and S2 project the endpoint estimate at every reverse step without source anchoring; S2 incorporates projection into the training loss, whereas S1 does not.
S3 anchors the source only at sampling time on a backbone trained on unconstrained noise.
Across all datasets, S1 and S2 achieve  $\mathrm{CE}$ comparable to PCFM, ECI, and CAFM, but show little improvement in $W_2$ over the unconstrained backbone.
Endpoint projection alone therefore enforces admissibility without resolving distributional bias, while incorporating it into training provides no measurable improvement.
S3 yields results similar to FFM and PBFM, with the highest  $\mathrm{CE}$ among the ablated variants and no improvement in $W_2$, reflecting the mismatch between the unconstrained training source and the anchored sampling source.
Neither component alone reproduces the improvements achieved by SAPC, indicating that reducing both  $\mathrm{CE}$ and $W_2$ requires combining source anchoring with a training objective that accounts for the anchored source distribution. 
Full ablation results are provided in Appendix~\ref{app:ablation}.

\begin{figure}[!h]
\centering
\includegraphics[width=\linewidth]{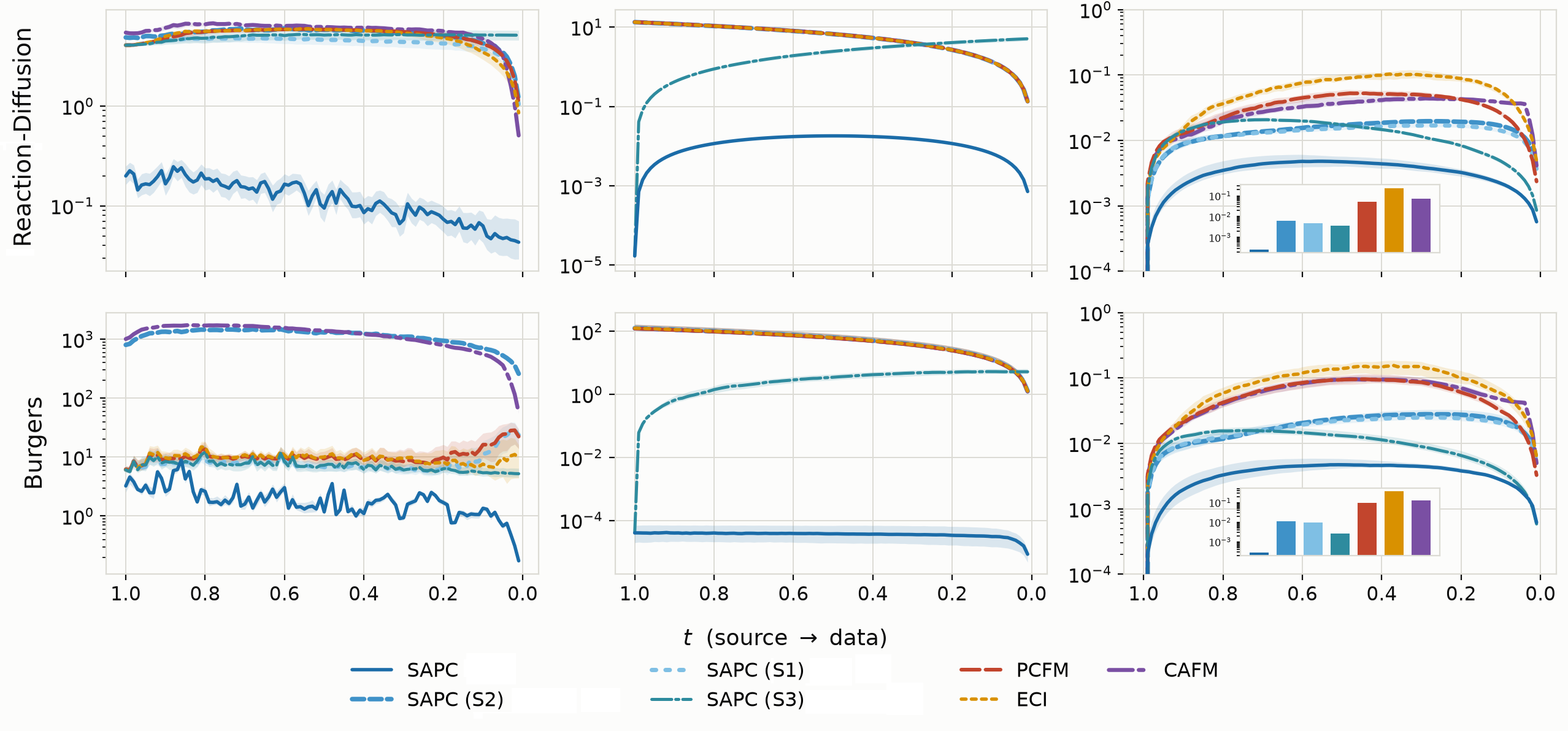}
\caption{Trajectory analysis (non-linear RD, top, and linear Burgers, bottom). 
\textbf{Left:} residual of endpoint estimate, $\|\mathcal{R}(\mathcal{V}_{0,\theta})\|_2$, before projection. 
\textbf{Middle:} residual $\|\mathcal{R}(\mathcal{V}_t)\|_2$ till last reverse step. 
\textbf{Right:} deviation $\Delta$ of the trajectory from the FM interpolant; the inset bars report $\mathrm{KE}$. }
\label{fig:trajectory_analysis}
\end{figure}

\noindent \textbf{Trajectory analysis.}
To examine how source anchoring and endpoint projection jointly influence generation in SAPC, we track three quantities at each sampling step.
First, we assess whether the network predicts an admissible target by computing the constraint residual of the posterior-mean estimate before projection as a function of time, $|\mathcal{R}(\mathcal{V}_{0,\theta})|_2$. 
Second, we measure how strongly each iterate violates the constraints defining $\mathcal{H}$, through the residual $|\mathcal{R}(\mathcal{V}_t)|_2$. 
Third, we quantify the deviation of the flow from the interpolant connecting the source $\mathcal{V}_1$ to the target $\mathcal{V}_0$ as $\Delta = |\mathcal{V}_t-[(1-t)\mathcal{V}_0+t\mathcal{V}_1]|_2/|\mathcal{V}_0-\mathcal{V}_1|_2$, normalising by the source–target distance to enable comparison across datasets.
As an integral measure, we also report the excess kinetic energy over the full sampling trajectory, defined as the difference between $\mathrm{KE} = \sum_t |\mathcal{V}_{t} - \mathcal{V}_{t-\Delta t}|_2^2 / \Delta t$ and the interpolant energy $\mathrm{KE}_{\mathrm{int}} = |\mathcal{V}_0 - \mathcal{V}_1|_2^2$.
Figure~\ref{fig:trajectory_analysis} compares SAPC, its ablated variants (S1–S3), and the projection-based competitors on RD (top) and Burgers (bottom). 
The curves show the median across generated samples at each sampling step, using a single random seed and data split. 
The left column shows that SAPC reduces the constraint residual of $\mathcal{V}_{0,\theta}$ by approximately one order of magnitude on RD and two on Burgers relative to unanchored methods throughout sampling. 
This indicates that the network learns to predict nearly admissible states, leaving only small corrections to the posterior-mean estimate during inference.
The middle column provides empirical support for Equation~\ref{eq:sapc-nonlinear-interp}. 
On Burgers, where $\mathcal{R}$ is affine, SAPC maintains a constraint residual on the order of $10^{-5}$–$10^{-4}$ throughout sampling.
The iterates $\mathcal{V}_t$ therefore lie on the constraint manifold, being those residuals rounding errors of the sampling loop rather than departures from $\mathcal{H}$.
On RD, the residual follows the profile predicted by Equation~\ref{eq:sapc-nonlinear-interp}, peaking near $t=1/2$ and decreasing symmetrically towards both endpoints.
Finally, the right column shows that SAPC reduces excess $\mathrm{KE}$ by two to three orders of magnitude relative to competing methods. 
On both RD and Burgers, the sampling trajectory remains close to the interpolant connecting the source and target endpoints, whereas competing methods deviate further from it and rely on the sampling-time projector to correct this effect.

\section{Conclusion}
We introduced Source Anchoring for Physical Consistency (SAPC), a Functional Flow Matching method that projects the source of the flow onto the manifold of physically admissible states, improving distributional fidelity over projection-based, gradient-based, and soft-constraint competitors by one to three orders of magnitude on Heat, Reaction–Diffusion and Burgers, and by a smaller margin on the Stokes tasks and Navier–Stokes, while matching the constraint precision of the best projection-based baselines. 
The ablation identifies the pairing of source projection with a matched training objective as the mechanism driving these distributional gains. 
Anchoring the source keeps the flow close to the manifold, so generation refines a near-consistent sample rather than reshaping a noisy one, avoiding the large corrections that projection-based methods apply to restore admissibility and that can drive samples onto admissible but off-distribution states.

\subsection*{AI use statement}

In this work, we used generative AI tools for create or modify scientific figures or images, create or edit software code, and edit a research paper to improve readability. 
We have not used generative AI tools for help develop theoretical models or conceptual frameworks, formulate mathematical claims, provide critical ingredients for proving mathematical claims, assist in the writing of proofs, propose or refine hypotheses, implement methods, design or provide feedback on research methodology or experiments, interpret results and support qualitative and thematic data analysis. 
Assist with translation, generate synthetic data sets or clean and reformat dataset are not applicable to this work. 
We have reviewed all AI-assisted work. 
We take responsibility for the final content of this work, including text, claims or artifacts produced with the aid of generative AI.

\subsection*{Reproducibility statement}
Code for data generation, training, and sampling of SAPC, its ablation variants, and all baselines is available at~\url{https://anonymous.4open.science/r/SAPC}.

\bibliography{main}
\bibliographystyle{iclr2027_conference}

\newpage
\appendix
\textbf{Supplementary Material}

\section{SAPC constraining}
\subsection{Functional Flow Matching}
\label{app:ffm}

Flow Matching (FM)~\citep{lipman2023} is a generative framework that learns a time-dependent vector field $u_{t}: \mathbb{R}^d \times [0,1] \to \mathbb{R}^d$ defining a continuous time-dependent diffeomorphism called the flow $\psi_{t}: \mathbb{R}^d \times [0,1] \to \mathbb{R}^d$ via the following ODE:
\begin{equation}
\frac{\mathrm{d}}{\mathrm{d}t}\, \psi_{t}(x) \;=\; u_{t}(\psi_{t}(x)), \qquad t \in [0, 1]\
\label{eq:fm-ode}
\end{equation}
that transports a source noise distribution $p_{1}$ at $t = 1$ to the target distribution $p_0$ at $t = 0$ along a prescribed path of marginals $p_{t}$ satisfying the continuity equation $\partial_{t} p_{t} + \nabla \cdot (p_{t} u_{t}) = 0$.
The vanilla FM objective regresses a learnable vector field $v_{\theta}$ onto the ground truth marginal vector field $u_{t}$ that generates the path $p_{t}$:
\begin{equation}
\mathcal{L}_{\mathrm{FM}} \;=\; \mathbb{E}_{t, x_{t} \sim p_{t}(x)} \left[ \left\| v_{\theta}(x_{t}, t) - u_{t}(x_{t}) \right\|^{2} \right]
\label{eq:fm-loss-vanilla}
\end{equation}
Generally $u_{t}$ is not available in closed form, making equation~\ref{eq:fm-loss-vanilla} intractable.
When conditioned on a pair of endpoints $(x_{0}, x_{1})$, however, the optimal-transport path is the linear interpolant $x_{t} = (1 - t) x_{1} + t x_{0}$, along which the conditional vector field is constant, $u_{t}(x_{t} \mid x_{0}, x_{1}) = x_{0} - x_{1}$. Regressing $v_{\theta}$ onto this conditional velocity yields the same gradients in $\theta$ as equation~\ref{eq:fm-loss-vanilla}~\citep{lipman2023}, giving the tractable conditional FM training objective:
\begin{equation}
\mathcal{L}_{\mathrm{CFM}} \;=\; \mathbb{E}_{t, x_{0}, x_{1}} \left[ \left\| v_{\theta}(x_{t}, t) - (x_{0} - x_{1}) \right\|^{2} \right]
\label{eq:fm-loss}
\end{equation}
with $t \sim \mathcal{U}[0,1]$, $x_{0} \sim p_{0}$, and $x_{1} \sim p_{1}$. Generation reduces to integrating equation~\ref{eq:fm-ode} backward in $t$ from a source draw $x_1 \sim p_1$ using the learned vector field $v_{\theta}$.

Functional Flow Matching (FFM)~\citep{kerrigan2023} extends FM to function space, so that the transported objects are the PDE solutions $\mathcal{V}: \Omega \to \mathbb{R}$, with $\Omega$ the spatio-temporal domain of the system.
Specifically, FFM learns a time-dependent vector field operator $v_{t}: \mathcal{U} \times [0,1] \to \mathcal{U}$ over a real separable Hilbert space $\mathcal{U}$ of functions $\mathcal{V}$, that defines a time-dependent diffeomorphism $\psi_{t}: \mathcal{U} \times [0,1] \to \mathcal{U}$, called the flow, via the ODE:
\begin{equation}
\partial_{t} \psi_{t}(\mathcal{V}) \;=\; v_{t}(\psi_{t}(\mathcal{V})), \qquad t \in [0, 1]
\label{eq:ffm-ode}
\end{equation}
The source noise distribution $p_1$ is a fixed noise measure over $\mathcal{U}$, induced by a Gaussian process with trace-class covariance operator $C$ built from a Mat\'{e}rn kernel.
The flow transports $p_1$ at $t = 1$ to the target state distribution $p_0(\mathcal{V})$ at $t = 0$.
As in FM, the vanilla FFM objective is intractable.
Under regularity conditions on $p_{0}$ and $p_{1}$~\citep{kerrigan2023}, the same conditional construction yields a tractable objective along the optimal-transport path of measures, where the ground truth states are linearly interpolated with source draws as $\mathcal{V}_{t} = (1 - t) \mathcal{V}_{1} + t \mathcal{V}_{0}$:
\begin{equation}
\mathcal{L}_{\mathrm{FFM}} \;=\; \mathbb{E}_{t, \mathcal{V}_{0}, \mathcal{V}_{1}} \left[ \left\| v_{\theta}(\mathcal{V}_{t}, t) - (\mathcal{V}_{0} - \mathcal{V}_{1}) \right\|^{2} \right]
\label{eq:ffm-loss}
\end{equation}
with $t \sim \mathcal{U}[0,1]$, $\mathcal{V}_{0} \sim p_{0}$, $\mathcal{V}_{1} \sim p_{1}$, and the norm the standard $L^{2}(\Omega)$ norm for square-integrable functions.
Generation reduces to integrating equation~\ref{eq:ffm-ode} backward in $t$ from a source draw $\mathcal{V}_1 \sim p_{1}$ using the learned vector field operator $v_{\theta}$.

\subsection{Projection}
\label{app:projector}

For each benchmark, $\mathcal{H}$ embeds two constraints: a
\textit{local} constraint pinning IC/BC (or initial vorticity for Navier-Stokes); and a \textit{global} constraint enforcing an integral conservation law along
the whole trajectory. 
Their residuals are concatenated into a single
$\mathcal{R}(\mathcal{V}) = [\mathcal{R}_\text{loc}(\mathcal{V}),\, \mathcal{R}_\text{glob}(\mathcal{V})]$.
The projector $P$ realizing the metric projection onto $\mathcal{H}$ then takes
one of two following forms, depending on whether the stacked residual $\mathcal{R}(\mathcal{V})$ is affine.

\textbf{Linear form.} When both blocks $\mathcal{R}_\text{loc}(\mathcal{V})$ and $\mathcal{R}_\text{glob}(\mathcal{V})$ are affine, $\mathcal{R}(\mathcal{V}) =
A\mathcal{V} - b$, so
$P$ reduces to the orthogonal projector:
\begin{equation}
    P(\mathcal{V}) \;=\; \mathcal{V} - A^\top (A A^\top)^{-1}(A\mathcal{V}-b)
\end{equation}
This is the case for the Heat, Burgers, and Navier-Stokes datasets.
The Gram matrix $AA^\top$ is factorized once per benchmark via a Cholesky decomposition, cached, and reused across all projections. 
We fall back to a pivoted LU factorization when Cholesky fails, and to a
pseudo-inverse when $AA^\top$ is rank-deficient. 
The factorisation and solve are carried out in \texttt{float64}.

\textbf{Non-linear form.} If either block $\mathcal{R}_\text{loc}(\mathcal{V})$ or $\mathcal{R}_\text{glob}(\mathcal{V})$ is non-linear, the concatenated residual $\mathcal{R}(\mathcal{V})$ is projected by Gauss-Newton. 
Starting from the pre-projection point $\mathcal{V}_0$, the update:
\begin{equation}
\mathcal{V}_{k+1} \;=\; \mathcal{V}_0 \;-\; J_k^\top
    \bigl(J_k J_k^\top + \mu I\bigr)^{-1}
    \bigl(h(\mathcal{V}_k) - J_k(\mathcal{V}_k - \mathcal{V}_0)\bigr),
\qquad J_k = \nabla \mathcal{R}(\mathcal{V}_k)
\label{eq:gauss-newton}
\end{equation}
is iterated for at most $n_\text{iter}$ steps, with Tikhonov damping
$\mu = 10^{-8}$ and early stopping when $\|\mathcal{R}(\mathcal{V}_k)\|_\infty < 10^{-6}$.
We set $n_\text{iter}=3$ on Reaction-Diffusion and $n_\text{iter}=10$ on
Stokes. 
This implementation follows the one reported by \citet{utkarsh2025}.
A single iteration ($n_\text{iter}=1$) recovers the tangent-space linearised projection at $\mathcal{V}_0$ used by ECI~\citep{cheng2025}.

\subsection{Residual bounds for source anchoring}
\label{app:residual-bounds}

\textbf{Residual along the interpolant.}
Let $\mathcal{V}_{0}, \bar{\mathcal{V}}_{1} \in \mathcal{H}$, so that $\mathcal{R}(\mathcal{V}_{0}) = \mathcal{R}(\bar{\mathcal{V}}_{1}) = 0$, and let $\mathcal{V}_{t} = (1-t)\mathcal{V}_{0} + t\bar{\mathcal{V}}_{1}$ for $t \in [0,1]$.
When $\mathcal{R}$ is affine, $\mathcal{R}(\mathcal{V}) = A\mathcal{V} - b$, and linearity gives:
\begin{equation}
\mathcal{R}(\mathcal{V}_{t})
= (1-t)\,\mathcal{R}(\mathcal{V}_{0}) + t\,\mathcal{R}(\bar{\mathcal{V}}_{1})
= 0, \qquad \forall\, t \in [0,1]
\label{eq:app-sapc-affine-interp}
\end{equation}
so the entire flow trajectory lies on $\mathcal{H}$.
When $\mathcal{R}$ is $C^{2}$ and non-linear, we define $\Delta = \bar{\mathcal{V}}_{1} - \mathcal{V}_{0}$ and $\phi(t) = \mathcal{R}(\mathcal{V}_{t}) = \mathcal{R}(\mathcal{V}_{0} + t\Delta)$, so that $\phi(0) = \phi(1) = 0$. 
Differentiating twice along the segment yields\footnote{The constant $L$ is finite because the segment
$[\mathcal{V}_{0}, \bar{\mathcal{V}}_{1}] =
\{\mathcal{V}_{0} + t\Delta : t \in [0,1]\}$ is compact in $L^{2}(\Omega)$
and $\nabla^{2}\mathcal{R}$ is continuous on it by the $C^{2}$ assumption, so the upper limit exists.}:
\begin{equation}
\phi''(t)
\;=\; \Delta^{\top}\, \nabla^{2}\mathcal{R}(\mathcal{V}_{t})\, \Delta,
\qquad
\|\phi''(t)\| \;\leq\; L\,\|\Delta\|^{2},
\qquad
L \;=\; \sup_{\mathcal{V}\in[\mathcal{V}_{0},\bar{\mathcal{V}}_{1}]}
\bigl\|\nabla^{2}\mathcal{R}(\mathcal{V})\bigr\|
\label{eq:app-sapc-phi-second}
\end{equation}
Taylor's formula with integral remainder at $t=0$ is $\phi(t) = t\,\phi'(0) + \int_{0}^{t}(t-s)\,\phi''(s)\,ds$. 
Evaluating at $t = 1$ and using $\phi(1) = 0$ gives $\phi'(0) = -\!\int_{0}^{1}(1-s)\,\phi''(s)\,ds$, that substituted back gives the Green's-function representation:
\begin{equation}
\phi(t)
\;=\; -\int_{0}^{1} G(t,s)\, \phi''(s)\, ds,
\qquad
G(t,s) \;=\;
\begin{cases}
s(1-t), & 0 \leq s \leq t\\
t(1-s), & t \leq s \leq 1
\end{cases}
\label{eq:app-sapc-green}
\end{equation}
of the two-point boundary problem $-\phi'' = f$ with homogeneous Dirichlet
endpoints. Combining $\|\phi''(t)\| \leq L\|\Delta\|^{2}$ with
$\int_{0}^{1} G(t,s)\, ds = \tfrac{1}{2} t(1-t)$ yields
\begin{equation}
\bigl\|\mathcal{R}(\mathcal{V}_{t})\bigr\|
\;\leq\; \tfrac{1}{2}\, t(1-t)\, L\,
\bigl\|\mathcal{V}_{0} - \bar{\mathcal{V}}_{1}\bigr\|^{2}
\label{eq:app-sapc-nonlinear-interp}
\end{equation}
which is the bound stated in~\eqref{eq:sapc-nonlinear-interp}.

\textbf{Bounding the interpolant marginals.}
From equation~\ref{eq:app-sapc-nonlinear-interp}, a line $(1-t)\mathcal{V}_{0} + t\bar{\mathcal{V}}_{1}$ drifts off $\mathcal{H}$ by at most $\tfrac{1}{2} t(1-t) L \|\mathcal{V}_{0} - \bar{\mathcal{V}}_{1}\|^{2}$.
The claim we make in the main text is about the interpolant marginals $p_{t}$, i.e.\ the distributions of $\mathcal{V}_{t} = (1-t)\mathcal{V}_{0} + t\bar{\mathcal{V}}_{1}$ as $t$ varies over $[0,1]$. 
Bridging the two requires taking expectations over $(\mathcal{V}_{0}, \bar{\mathcal{V}}_{1})$.
The endpoint gap has finite second moment. 
Indeed,
\begin{equation}
\mathbb{E}\bigl\|\mathcal{V}_{0} - \bar{\mathcal{V}}_{1}\bigr\|^{2}
\;\leq\; 2\bigl(\mathbb{E}\|\mathcal{V}_{0}\|^{2}
              + \mathbb{E}\|\bar{\mathcal{V}}_{1}\|^{2}\bigr)
\;=:\; M \;<\; \infty
\label{eq:app-sapc-second-moment}
\end{equation}
where $\mathbb{E}\|\mathcal{V}_{0}\|^{2} < \infty$ holds by assumption on $p_{\mathcal{V}}$ (because the solutions of all PDE datasets considered in this study are bounded on the compact spatio-temporal domain $\Omega$), and $\mathbb{E}\|\bar{\mathcal{V}}_{1}\|^{2} < \infty$ is inherited from $\mathcal{GP}(0, C)$ via the projection\footnote{$P$ preserves finite second moments: 1-Lipschitz in the affine case, locally Lipschitz in the $C^{2}$ non-linear case by the implicit function theorem applied to~\eqref{eq:gauss-newton}. 
Together with $\mathbb{E}\|\mathcal{V}_{1}\|^{2} = \mathrm{tr}(C) < \infty$ this gives $\mathbb{E}\|\bar{\mathcal{V}}_{1}\|^{2} < \infty$.}. 
Taking expectations of~\eqref{eq:app-sapc-nonlinear-interp} therefore yields the marginal residual bound
\begin{equation}
\mathbb{E}_{\mathcal{V}_{t} \sim p_{t}}
\bigl\|\mathcal{R}(\mathcal{V}_{t})\bigr\|
\;\leq\; \tfrac{1}{2}\, t(1-t)\, L\, M
\label{eq:app-sapc-marginal-bound}
\end{equation}
From Markov's inequality it follows, for every $\varepsilon > 0$:
\begin{equation}
\Pr\bigl(\|\mathcal{R}(\mathcal{V}_{t})\| \leq \varepsilon\bigr)
= 1 - \Pr\bigl(\|\mathcal{R}(\mathcal{V}_{t})\| > \varepsilon\bigr)
\geq 1 - \frac{t(1-t)\, L\, M}{2\,\varepsilon}
\;\xrightarrow[t \to 0,\, 1]{}\; 1
\label{eq:app-sapc-markov}
\end{equation}
Most samples drawn from $p_{t}$ therefore have a residual $\|\mathcal{R}(\mathcal{V}_{t})\|$ below the threshold $\varepsilon$, with only a small fraction violating it; that fraction shrinks as $t$ approaches either endpoint of the flow.

\section{Dataset generation and definition of constraint manifolds}
\label{app:datasets}

This section provides a detailed description of the datasets used in this work and their generation. 
To monitor sensitivity to the constraint enforced at test time, we generate, for each dataset, four held-out test sets of $1225$ trajectories each, differing only in the quantity held fixed within the set: the phase $\phi$ of the IC on Heat; the IC on RD and on Burgers; the initial vorticity on NS; the decay rate $k$ on Stokes IC, and the oscillation frequency $\omega$ on Stokes BC. 
The remaining parameter is redrawn independently per trajectory within the set (the diffusivity $\alpha$ on Heat, the viscosity $\nu$ on Burgers, the forcing phase on NS, $\omega$ on Stokes IC and $k$ on Stokes BC) so each test set is a distribution of solutions under a single fixed constraint rather than a single trajectory.
See the corresponding subsections for more details. 
Each split (training, validation, and the four held-out test sets) is drawn from an independent random stream.

\subsection{Heat Equation}

We consider the one-dimensional heat equation with periodic boundary conditions:
\begin{equation}\label{eq:heat}
    \frac{\partial u(x,t)}{\partial t} = \alpha \, \frac{\partial^2 u(x,t)}{\partial x^2},
    \quad x \in [0, 2\pi], \quad t \in [0,1]
\end{equation}
\begin{equation}\label{eq:heat_ic}
    u(x,0) = \sin(x+\phi), \quad u(0,t) = u(2\pi, t)
\end{equation}
\noindent The equation describes the distribution of heat $u(x,t)$ along a ring with non-uniform initial temperature $u(x,0)$. As time passes, the heat redistributes and smooths out by internal conduction, with closed-form solution $u(x,t) = e^{-\alpha t}\sin(x+\phi)$.

\begin{figure}[!h]
\begin{center}
\includegraphics[width=\textwidth]{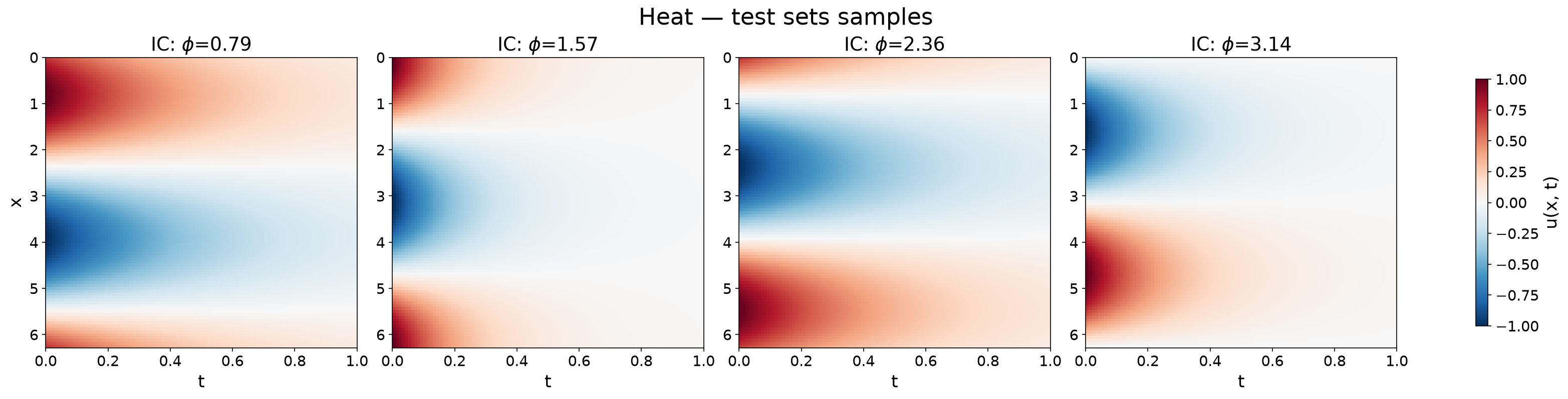}
\end{center}
\caption{Representative test-set trajectories of the heat benchmark, one per held-out IC. Each panel is drawn from a different test split, pinned to a distinct initial phase $\phi \in \{\pi/4, \pi/2, 3\pi/4, \pi\}$, while the diffusivity $\alpha$ is drawn from the training range $[1,5]$ within each split. The heat solution $u(x,t) = e^{-\alpha t}\sin(x+\phi)$ decays exponentially in time while preserving the spatial profile set by $\phi$.}
\label{fig:heat_test_samples}
\end{figure}

To train and validate the models, we generate $6400$ and $1225$ train and validation
trajectories respectively, with the diffusion coefficient and phase sampled uniformly
and independently for each trajectory, $\alpha \sim U[1,5]$ and $\phi \sim U[0,\pi]$,
following the same setting as~\citep{cheng2025, utkarsh2025}.
To evaluate the models, we generate four held-out test sets of $1225$ trajectories each.
Within a test set the phase is fixed to one of $\phi \in \{\pi/4,\, \pi/2,\, 3\pi/4,\, \pi\}$,
so that all of its trajectories share the same initial condition
$u_{\text{IC}}(x,0) = \sin(x + \phi)$, while the diffusivity $\alpha \sim U[1, 5]$ is redrawn
independently for each trajectory. Each test set is therefore a distribution over
solutions under a \textbf{fixed IC}, and the
test sets differ only in which IC is held fixed (Figure~\ref{fig:heat_test_samples}).

The projector is defined to map the diffusion trajectory onto the manifold of states
consistent with the fixed IC and the \textbf{Linear Mass Conservation (LMC)} of the
system:
\begin{equation}\label{eq:heat_H}
    \mathcal{H} = \left\{ u : [0, 2\pi] \times [0,1] \to \mathbb{R}
    \ \middle|
    \begin{array}{l}
        u(x,0) = u_{\text{IC}}(x), \ \ \forall\, x \in [0, 2\pi]\,; \\[2pt]
        \int_0^{2\pi} u(x,T)\,dx - \int_0^{2\pi} u(x,0)\,dx = 0 ,\ \ \forall\, T \in [0,1]
    \end{array}
    \right\}
\end{equation}
\noindent being $m(t) = \int_0^{2\pi} u(x,t)\,dx$ the mass of the system.

\subsection{Reaction-Diffusion (RD) Equation}

We consider the non-linear one-dimensional reaction-diffusion equation with Neumann boundary conditions:
\begin{equation}\label{eq:rd}
    \frac{\partial u(x,t)}{\partial t} = \nu \, \frac{\partial^2 u(x,t)}{\partial x^2} + \rho \, u(x,t)\big(1 - u(x,t)\big),
    \quad x \in [0, 1], \quad t \in [0,1]
\end{equation}
\begin{equation}\label{eq:rd_bc}
    -\nu\left.\frac{\partial u(x,t)}{\partial x}\right|_{x=0} = g_L, \qquad
    -\nu\left.\frac{\partial u(x,t)}{\partial x}\right|_{x=1} = g_R
\end{equation}
\noindent The equation describes the evolution of a population density (or concentration) $u(x,t)$ diffusing along a 1D strip with diffusivity $\nu$, while growing locally according to logistic reaction kinetics $\rho \, u(1 - u)$ with growth rate $\rho$, saturating at the carrying capacity $u=1$. We set $\nu=0.005$ and $\rho=0.01$, as in~\citep{utkarsh2025}. 
The boundary fluxes at the left and right ends of the domain are specified as $g_L$ and $g_R$, respectively.
The initial condition $u_{\text{IC}}(x,0)$ is sampled from a randomized combination of sinusoidal and localized bump functions. 
Specifically, each initial condition is drawn as a random superposition of up to three sinusoidal modes, with amplitudes and phases sampled uniformly. 
With probability $0.1$ the profile is additionally localized by a smooth $\tanh$ window,
producing bump-like states. Every realization is rescaled to $[0,1]$ range.

\begin{figure}[!h]
\begin{center}
\includegraphics[width=\textwidth]{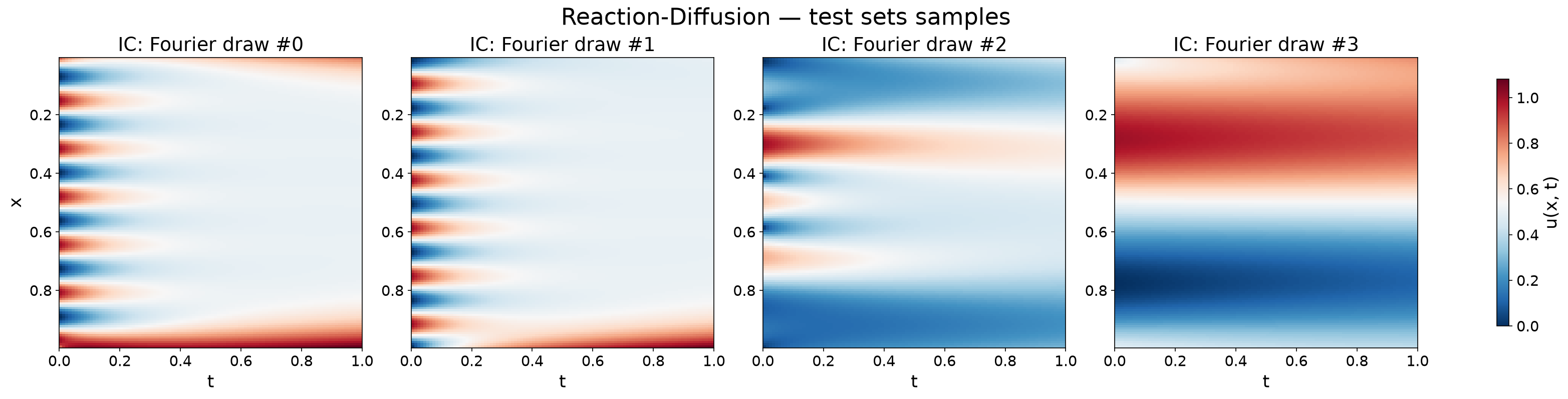}
\end{center}
\caption{Representative test-set trajectories of the RD benchmark, one per held-out IC.  Each panel is drawn from a different test split, pinned to a distinct random low-mode Fourier draw $u_{\text{IC}}(x,0)$, while the Neumann boundary fluxes $(g_L, g_R)$ vary per sample, drawn from the same training ranges within each split. The solution's dynamics are dominated by diffusive smoothing of the initial fine-scale Fourier structure and by mass injected or removed through the Neumann boundary fluxes.}
\label{fig:rd_test_samples}
\end{figure}

To train and validate the models, we generate $6400$ and $1225$ train and
validation trajectories respectively, by combining $80$ (or $35$)
initial conditions $u_{\text{IC}}$ with $80$ (or $35$) boundary conditions, with
$(g_L, g_R)$ sampled uniformly, $g_L \sim U[0, 0.05]$ and $g_R \sim U[-0.05, 0]$.
To evaluate the models, we generate four held-out test sets of $1225$ trajectories
each. Every test set pins a $u_{\text{IC}}$, drawn from the same distribution, and pairs it with $1225$ independent boundary-flux draws.
Each test set is therefore a distribution over solutions sharing one \textbf{fixed IC}, and the
test sets differ only in which IC is held fixed (Figure~\ref{fig:rd_test_samples}).

The projector is defined to map the diffusion trajectory onto the manifold of states
consistent with the fixed IC and the \textbf{Non-Linear Mass Conservation (NLMC)} of
the system:
\begin{equation}\label{eq:rd_H}
    \mathcal{H} = \left\{ u : [0,1] \times [0,1] \to \mathbb{R}
    \ \middle|
    \begin{array}{l}
        u(x,0) = u_{\text{IC}}(x), \ \ \forall\, x \in [0,1]\,; \\[2pt]
        \int_0^{1} \big[u(x,T) - u(x,0)\big]\,dx = \int_0^{T}\big(g_L - g_R\big)\,ds \,+ \\[2pt]
        \quad + \rho\int_0^{T}\!\!\int_0^{1} u(x,s)\big(1-u(x,s)\big)\,dx\,ds ,
          \ \ \forall\, T \in [0,1]
    \end{array}
    \right\}
\end{equation}
\noindent being $m(t) = \int_0^1 u(x,t)\,dx$ the mass of the system.
The constraint~\eqref{eq:rd_H} involves the boundary fluxes $(g_L,g_R)$, which are
unknown for a field produced by the generative model. At evaluation time we therefore
estimate them from the field itself by one-sided finite differences of the Neumann
condition, $g_L = -\nu\,\partial_x u|_{x=0} \approx -\nu\,(u_1 - u_0)/\Delta x$ and
$g_R = -\nu\,\partial_x u|_{x=1} \approx -\nu\,(u_{N-1} - u_{N-2})/\Delta x$. This
estimator is first-order accurate and evaluated one cell away from the boundary, so
the resulting residual retains an $O(\Delta x)$ bias: applied to the exact solutions
of the test sets it yields a non-zero constraint error, between
$1.7\times10^{-2}$ and $1.0\times10^{-1}$ depending on the test set. We therefore read
the NLMC constraint error as a \textit{relative} measure of mass-balance violation,
comparable across methods, rather than as an absolute deviation from the physical balance.

\subsection{Stokes Problem}

We consider Stokes' second problem, describing the unsteady flow of a viscous incompressible fluid generated by the oscillatory motion of a flat plate:
\begin{equation}\label{eq:stokes}
    \frac{\partial u(x,t)}{\partial t} = \nu \, \frac{\partial^2 u(x,t)}{\partial x^2},
    \quad x \in [0, 1], \quad t \in [0,1]
\end{equation}
\begin{equation}\label{eq:stokes_bc}
    u(x,0) = A e^{-kx} \cos(kx), \qquad u(0,t) = A \cos(\omega t)
\end{equation}
\noindent where $u(x,t)$ is the fluid velocity parallel to the plate, $x$ the distance from the plate, and $\nu$ the viscosity. The plate is set into motion at $t=0$ with sinusoidal velocity of amplitude $A=2$ and oscillation frequency $\omega$, with $k = \sqrt{\omega/(2\nu)}$.

\begin{figure}[!h]
\centering
\begin{subfigure}{\textwidth}
    \centering
    \includegraphics[width=\textwidth]{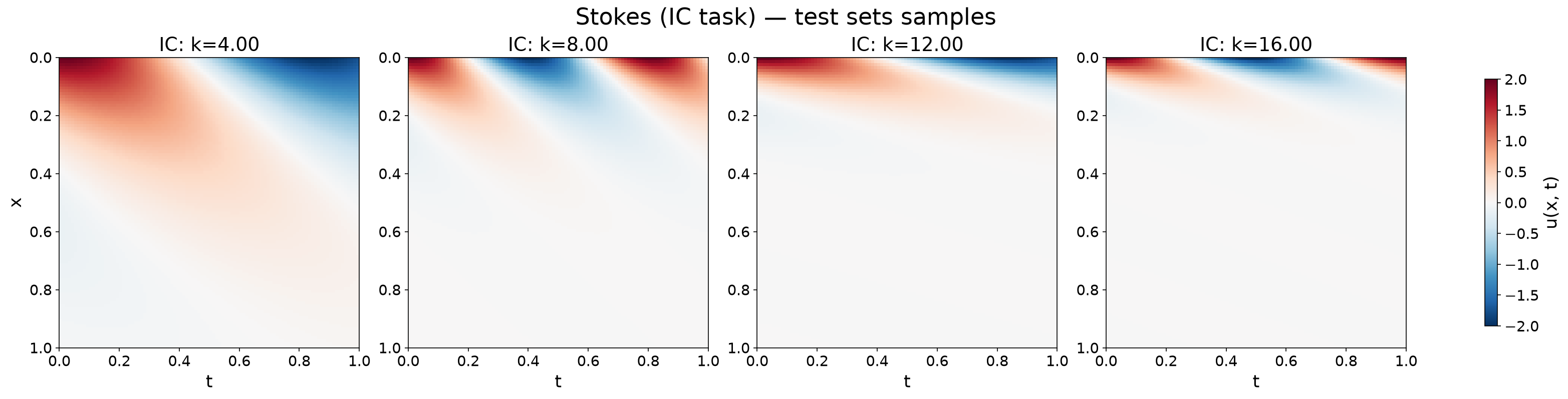}
\end{subfigure}
\vspace{0.5em}
\begin{subfigure}{\textwidth}
    \centering
    \includegraphics[width=\textwidth]{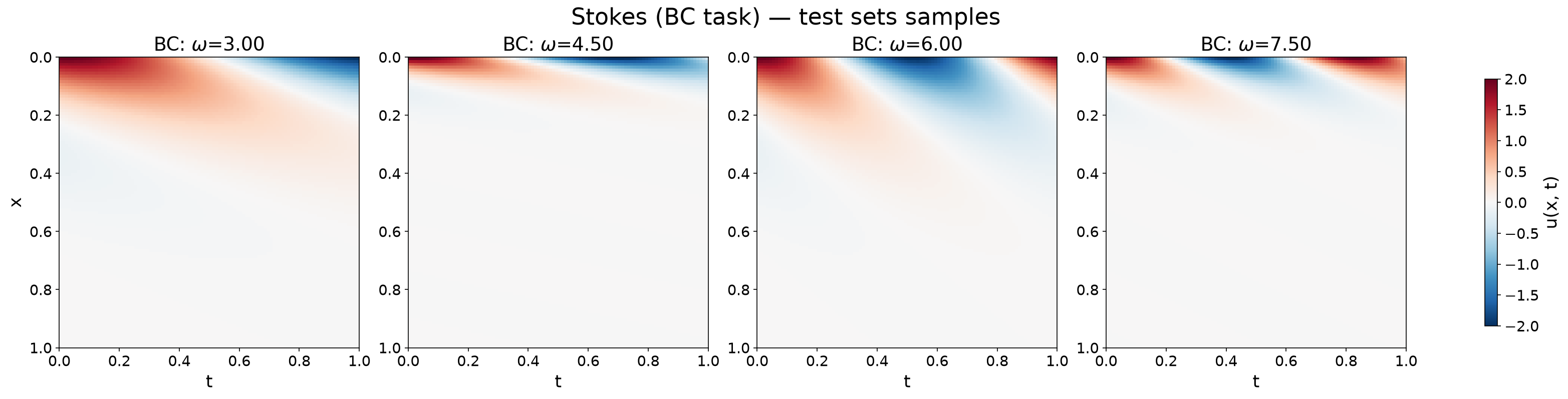}
\end{subfigure}
\caption{Representative test-set samples for the two Stokes tasks. Both solve the classic Stokes velocity field $u(x,t)$, a wave that decays exponentially away from the oscillating boundary at $x=0$, with decay rate and spatial wave number both set by $k$ and $\omega$. \textbf{IC task (top):} One sample from each of the four held-out test sets, each pinning a distinct decay rate $k \in \{4, 8, 12, 16\}$, while $\omega$ varies per sample, drawn from the training range $[2,8]$. Larger $k$ gives a thinner boundary layer and faster spatial oscillation near $x=0$. \textbf{BC task (bottom):} One sample from each of the four held-out test sets, each pinning a distinct frequency $\omega \in \{3, 4.5, 6, 7.5\}$, while $k$ varies within its training range $[2,20]$ per sample. Larger $\omega$ completes more oscillation periods over $t\in[0,1]$.}
\label{fig:stokes_test_samples}
\end{figure}

To train and validate the models, we generate $6400$ and $1225$ train and validation
trajectories respectively, by uniformly and independently sampling $\omega \sim U[2,8]$
and $k \sim U[2,20]$ for each trajectory, following the same setting
as~\citep{cheng2025}. 
The two parameters are sampled independently to decouple the
IC (governed by $k$) from the BC (governed by $\omega$);
the viscosity is then determined by $\nu = \omega / (2k^2)$. 
To evaluate the models, we consider two separate Stokes experiments: in
the \textbf{IC task}, $u_{\text{IC}}(x,0) = A e^{-kx}\cos(kx)$ is held out
through the decay rate $k$, while in the \textbf{BC task} the oscillating boundary
$u_{\text{BC}}(0,t) = A\cos(\omega t)$ is held out through the frequency $\omega$.
To this end, we generate four held-out test sets of $1225$ trajectories each
per task. 
In the IC task, each test set fixes $k \in \{4, 8, 12, 16\}$, so that all of
its trajectories share the same $u_{\text{IC}}(x,0)$, while
$\omega \sim U[2,8]$ is still redrawn per trajectory (Figure~\ref{fig:stokes_test_samples}, top). 
In the BC task, each test set
fixes $\omega \in \{3, 4.5, 6, 7.5\}$, so that all of its trajectories share the same $u_{\text{BC}}(0,t)$, while $k \sim U[2,20]$ is redrawn
per trajectory (Figure~\ref{fig:stokes_test_samples}, bottom). 
Each test set is therefore a distribution over solutions under a
single fixed constraint.

The projector is defined to map the diffusion trajectory onto the manifold of states
consistent with the constraint (either BC or IC) and the \textbf{Non-Linear Energy
Conservation (NLEC)} of the system. Multiplying \eqref{eq:stokes} by $u$ and
integrating over the spatial domain gives the energy balance:
\begin{equation}\label{eq:stokes_H}
    \mathcal{H} = \left\{ u : [0,1]\times[0,1] \to \mathbb{R}
    \ \middle|
    \begin{array}{l}
        u(x,0) = u_{\text{IC}}(x), \ \ \forall\, x \in [0,1] \quad \text{(for IC task)}\, \\[2pt]
        u(0,t) = A\cos(\omega t), \ \ \forall\, t \in [0,1] \quad \text{(for BC task)}\, \\[2pt]
        \int_0^1 u^2(x,T)\,dx - \int_0^1 u^2(x,0)\,dx = \\[2pt]
          2\nu \int_0^{T} \left[ u(1,t)\,\frac{\partial u(1,t)}{\partial x} 
          - u(0,t)\,\frac{\partial u(0,t)}{\partial x} \right] dt \, + \\[2pt]
         - \ 2\nu \int_0^{T}\!\!\int_0^{1} 
          \left(\frac{\partial u(x,t)}{\partial x}\right)^{2} dx \, dt,
          \ \ \forall\, T \in [0,1]
    \end{array}
    \right\}
\end{equation}
\noindent being $E(t) = \int_0^1 u^2(x,t)\,dx$ the energy of the
system, the first integral on the right side of the energy balance \eqref{eq:stokes_H} the net boundary
work done on the fluid, and the second the viscous dissipation.
Energy balance depends on the viscosity $\nu$, which is unknown for a generated field. 
During generation, we therefore estimate it per sample by least
squares on the governing \eqref{eq:stokes}, i.e.
$\hat{\nu} = \langle \frac{\partial u}{\partial t}, \frac{\partial^2 u}{\partial x^2}
\rangle / \langle \frac{\partial^2 u}{\partial x^2}, \frac{\partial^2 u}{\partial x^2}
\rangle$, with the spatial and temporal derivatives taken by finite differences. 
On the
ground-truth fields $\hat{\nu}$ recovers the true viscosity $\nu = \omega/(2k^2)$ to
within $4.7$-$5.4\%$ on the BC task and $3.7$-$16.4\%$ on the IC task, the latter
degrading with the held-out decay rate $k$ as the field concentrates into an
increasingly thin boundary layer. The residual of \eqref{eq:stokes_H} is accumulated
over the trajectory, so the first-order errors of the finite-difference derivatives do
not cancel: applied to the exact solutions it yields a constraint error between
$8.5\times10^{-2}$ and $1.1\times10^{-1}$ across the eight held-out test sets, rather
than zero. Substituting the exact $\nu$ in place of $\hat{\nu}$ raises it further, to
$1.27$-$1.40\times10^{-1}$, confirming that the accumulation, not the viscosity
estimate, sets this floor. The NLEC constraint error should therefore be read as a
comparison between methods, all scored with the same residual, and not as an absolute
measure of physical energy imbalance.

\subsection{Burgers' Equation}

We consider the one-dimensional viscous Burgers' equation with periodic boundary conditions:
\begin{equation}\label{eq:burgers}
    \frac{\partial u(x,t)}{\partial t} + u(x,t)\,\frac{\partial u(x,t)}{\partial x} = \nu \, \frac{\partial^2 u(x,t)}{\partial x^2},
    \quad x \in [0, 1], \quad t \in [0,1]
\end{equation}
\begin{equation}\label{eq:burgers_bc}
    u(x,0) = u_{\text{IC}}(x), \qquad u(0,t) = u(1, t)
\end{equation}
\noindent The equation combines non-linear advection, $u\, \frac{\partial u(x,t)}{\partial x}$, with linear diffusion of strength $\nu$, and is a canonical simplified version of the Navier-Stokes equations modelling the interplay between convection and viscosity. 
The initial condition $u_{\text{IC}}(x,0)$ is sampled from a randomized superposition of low-frequency Fourier modes.

To train and validate the unconstrained model, we generate 6400 and 1225 train
and validation trajectories respectively, with viscosity sampled uniformly,
$\nu \sim \mathcal{U}[0.01, 0.05]$, and initial condition realizations drawn
independently for each trajectory as randomized superpositions of four
low-frequency Fourier modes, rescaled to $[0,1]$ range.

\begin{figure}[!h]
\begin{center}
\includegraphics[width=\textwidth]{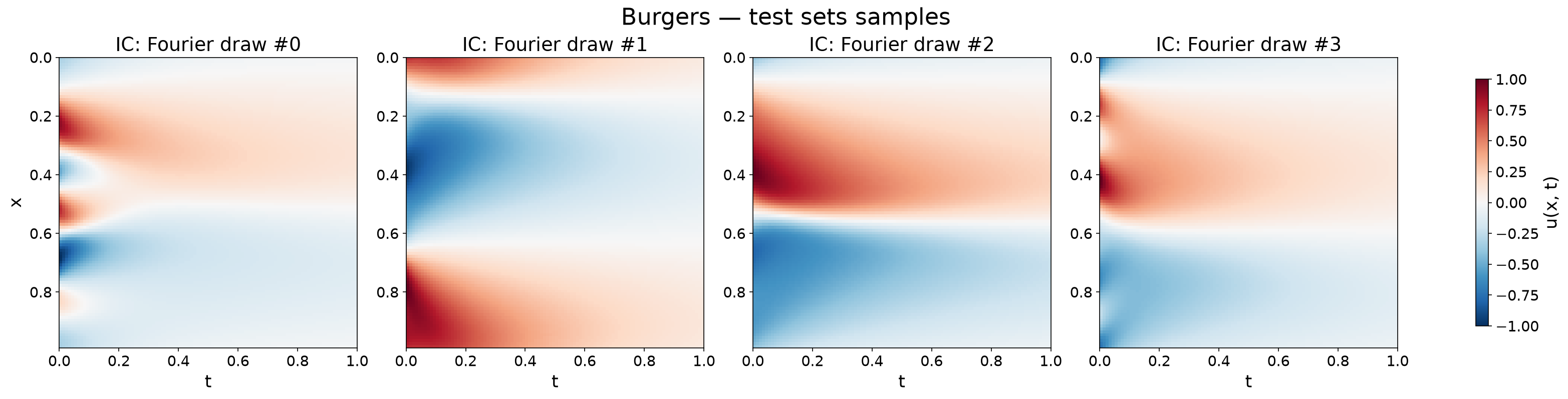}
\end{center}
\caption{Representative test-set trajectories of the Burgers benchmark. Each panel is drawn from a different test split, pinned to a distinct random low-mode Fourier draw $u_{\text{IC}}(x,0)$, while the viscosity $\nu$ varies per sample, drawn from the training range $[0.01, 0.05]$. Non-linear advection steepens the initial profile within the first few time steps, producing sharp layers whose thickness is set by $\nu$; the field then relaxes into near-uniform plateaus separated by these thin layers from then on.}
\label{fig:burgers_test_samples}
\end{figure}

To evaluate the models, we generate four held-out test sets of $1225$ trajectories
each. Every test set pins a $u_{\text{IC}}$, drawn from the same Fourier distribution, while the viscosity keeps varying over
$\mathcal{U}[0.01, 0.05]$ within the split.
Each test set is therefore a distribution over solutions sharing one \textbf{fixed IC}, and the test sets differ only in which IC is held fixed (Figure~\ref{fig:burgers_test_samples}).

The projector is defined to map the diffusion trajectory onto the manifold of states consistent with the fixed IC and the \textbf{Linear Mass Conservation (LMC)} of the system. Indeed, since the domain is periodic, the non-linear advective flux integrates to zero over $[0,1]$, so mass is exactly conserved:
\begin{equation}\label{eq:burgers_H}
    \mathcal{H} = \left\{ u : [0, 1] \times [0,1] \to \mathbb{R}
    \ \middle|
    \begin{aligned}
        &u(x,0) = u_{\text{IC}}(x), \ \ \forall\, x \in [0, 1] \\[2pt]
        &\int_0^{1} u(x,T)\,dx - \int_0^{1} u(x,0)\,dx = 0 ,\ \ \forall\, T \in [0,1]
    \end{aligned}
    \right\}
\end{equation}
\noindent being $m(t) = \int_0^1 u(x,t)\,dx$ the mass of the system.

\subsection{Navier-Stokes (NS) Equation}

We consider the two-dimensional Navier-Stokes equation for a viscous, incompressible fluid in vorticity form with periodic boundary conditions:
\begin{equation}\label{eq:ns}
    \frac{\partial w(\mathbf{x},t)}{\partial t} + \mathbf{u}(\mathbf{x},t) \cdot \nabla w(\mathbf{x},t) = \nu \, \Delta w(\mathbf{x},t) + f(\mathbf{x}),
    \quad \mathbf{x} \in [0,1]^2, \quad t \in [0,49]
\end{equation}
\begin{equation}\label{eq:ns_bc}
    \nabla \cdot \mathbf{u}(\mathbf{x},t) = 0, \qquad w(\mathbf{x},0) = w_0(\mathbf{x})
\end{equation}
\noindent where $\mathbf{u}(\mathbf{x},t)$ is the velocity field, $w = \nabla \times \mathbf{u}$ is the vorticity, $w_0$ is the initial vorticity, $\nu$ is the viscosity, and $f(\mathbf{x})$ is a fixed forcing term. 
The incompressibility constraint $\nabla \cdot \mathbf{u} = 0$ enforces conservation of mass, and the domain is treated with periodic boundary conditions in both spatial directions.

\begin{figure}[!h]
\begin{center}
\includegraphics[width=\textwidth]{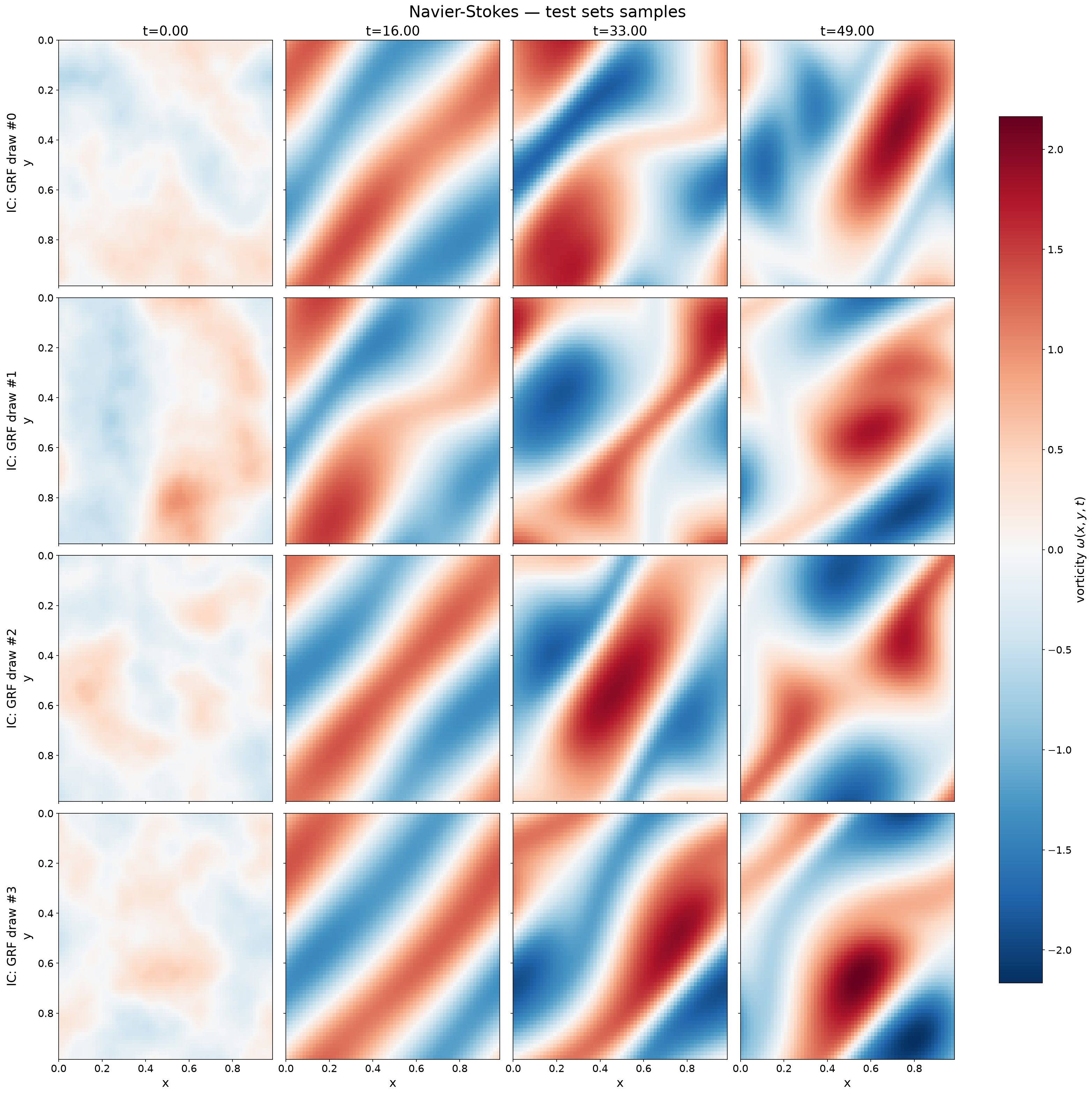}
\end{center}
\caption{One representative test-set trajectory of the 2D Navier-Stokes benchmark for each of its four held-out IC, shown as a $4\times4$ grid: each row is a different test split pinned to a distinct unseen initial vorticity field $\omega_0$ drawn from a Gaussian random field; each column is one of four evenly spaced snapshots across the rollout $t\in[0,49]$. The small-scale initial vorticity is rapidly overtaken by the large-scale pattern imposed by the diagonal forcing mode, with $\phi$ varying across trajectories.}
\label{fig:ns_test_samples}
\end{figure}

As in~\citep{utkarsh2025}, training and validation trajectories are obtained by pairing initial vorticities $w_0$, sampled from a Gaussian random field, with forcing realizations $f(\mathbf{x}) = 0.1\sqrt{2}\sin(2\pi(x_1+x_2)+\phi)$, $\phi \sim U[0,\pi/2]$, with viscosity fixed to $\nu = 10^{-3}$. Specifically, we draw 100 initial vorticities and 100 forcing phases and form the full $100 \times 100$ product for training (10000 trajectories), and likewise 35 initial vorticities and 35 forcing phases for validation ($35 \times 35 = 1225$ trajectories). 
To evaluate generalization to unseen initial conditions, we build four held-out test sets of 1225 trajectories each: every test set is generated from a single initial vorticity $w_0$, unseen during training, paired with 1225 independently sampled forcing phases $\phi \sim U[0,\pi/2]$; the four test sets thus differ only in their initial vorticity (Figure~\ref{fig:ns_test_samples}).

The projector is defined to map the diffusion trajectory onto the manifold of states consistent with the linear conservation of total vorticity of the system \textbf{(LVC, Linear Vorticity Conservation)}, induced by the periodicity of the domain and the divergence-free velocity field:
\begin{equation}\label{eq:ns_H}
    \mathcal{H} = \left\{ w : [0,1]^2 \times [0,49] \to \mathbb{R}
    \ \middle|
    \begin{aligned}
        &w(\mathbf{x},0) = w_0(\mathbf{x}), \ \ \forall\, \mathbf{x} \in [0,1]^2 \\[2pt]
        &\int_{[0,1]^2} w(\mathbf{x},T)\,d\mathbf{x} - \int_{[0,1]^2} w(\mathbf{x},0)\,d\mathbf{x} = \\[2pt]
        &\qquad = T \int_{[0,1]^2} f(\mathbf{x})\,d\mathbf{x} ,\ \ \forall\, T \in [0,49]
    \end{aligned}
    \right\}
\end{equation}
\noindent being $\Omega(t) = \int_{[0,1]^2} w(\mathbf{x},t)\,d\mathbf{x}$ the total vorticity of the system.

\section{Implementation details}

\subsection{Training and sampling setup}
\label{app:setup}

The velocity field $v_\theta(\mathcal{V}_{t_i}, t_i)$ is parametrized by a Fourier Neural Operator (FNO)~\citep{li2021}, following the setup of ECI~\citep{cheng2025} and PCFM~\citep{utkarsh2025}, and implemented in PyTorch~2.5.1.
The one-dimensional PDE systems are treated as two-dimensional fields over the space-time grid $(n_x, n_t)$ and are modelled by a 2D FNO, while Navier-Stokes is modelled by a 3D FNO over $(n_x, n_y, n_t)$.
The FNO input is the concatenation of the current state, a linear positional encoding with one channel for each axis of the field (the physical time included, $x,t$ in 2D; $x,y,t$ in 3D), and a sinusoidal embedding of the flow time.
Table~\ref{tab:fno-config} summarises the two backbone configurations.
Fields enter the network in raw units: no per-channel standardisation, min-max rescaling or normalisation is applied at any stage.
The source is drawn from a Gaussian Process with a Mat\'ern kernel (smoothness $\nu=0.5$, length scale $10^{-3}$, unit variance), sampled by a dense Cholesky factorisation on the $(n_x, n_t)$ grids and by circulant FFT synthesis on the larger Navier-Stokes volume, and the flow is integrated over $N=100$ steps.

\begin{table}[ht]
\centering
\caption{FNO backbone configurations. 2D FNO for 1D datasets; 3D FNO for 2D datasets.}
\label{tab:fno-config}
\begin{tabular}{lcc}
\toprule
 & 2D FNO  & 3D FNO  \\
\midrule
Fourier layers            & 4              & 4 \\
Fourier modes per axis    & 32             & 16 \\
Hidden channels           & 64             & 32 \\
Projection channels       & 256            & 128 \\
Time-embedding channels   & 32             & 16 \\
Training batch size & 256 & 32 \\
Sampling batch size & 32  & 16
\end{tabular}
\end{table}

The model is trained by minimising the FFM mean-squared-error objective
(Eq.~\ref{eq:ffm-loss}) with the Adam optimiser ($\beta_1=0.9$, $\beta_2=0.999$, no weight decay) at a base learning rate of $3\times10^{-4}$, for $2\times10^{4}$ optimiser steps, with the batch size of Table~\ref{tab:fno-config}: $256$ on the 2D backbone, reduced to $32$ on Navier-Stokes to fit the activation memory of the space-time volume.
The learning rate follows a reduce-on-plateau schedule (factor $0.5$, patience $10$ validation steps, floor $10^{-4}$) and gradients are clipped to a maximum norm of $100$.
Validation loss is evaluated every $200$ optimiser steps on the held-in validation split and is used to drive the schedule: no exponential moving average of
the weights and no validation-based checkpoint selection is performed, so every reported number comes from the final iterate at step $2\times10^{4}$. Training runs in \texttt{float32} throughout, with no mixed precision, under cuDNN deterministic kernels and per-run seeding of the PyTorch and CUDA generators.
SAPC additionally folds the projection of the training source into the loss as described in Section~\ref{sec:method}; all other hyperparameters are shared with the unconstrained backbone.
Each configuration is trained with three random seeds $\{0, 42, 205\}$.
Every run occupies one NVIDIA GH200 (96\,GB) with four data loader workers.
One seed of the unconstrained backbone takes $50$-$75$ minutes; SAPC costs the projection of the training source on top of that, from about $1$ hour on Heat and Burgers, where the projector is linear, to approximately $9$ hours on Navier-Stokes.

Generation integrates the reverse process backward in $t$ on the uniform grid $t_i = i/N$, $i = N,\dots,1$, with $N = 100$ as in the forward process. 
The FM objective draws $t \sim \mathcal{U}[0,1]$ continuously rather than on a grid, so there is no training-time discretisation for sampling to match.
Table~\ref{tab:projector-config} gives the sampling-time projector settings.
On the datasets whose constraint families are affine (Heat, Burgers, Navier-Stokes) the projection is the exact orthogonal projector $P = I - A^\top (A A^\top)^{-1} A$, applied in one solve through a Gram factorisation cached across steps. 
On RD and the two Stokes tasks the constraint is curved and $P$ is the damped Gauss-Newton iteration of Eq.~\ref{eq:gauss-newton}, with Tikhonov damping $\mu = 10^{-8}$, stopped early once $\lVert h(u_k)\rVert_\infty < 10^{-6}$. 
Stokes stacks a linear pin with the non-linear kinetic-energy balance and stalls at $\lVert h\rVert_\infty \approx 10^{-4}$ after three iterations, so its cap is raised to ten, at which the residual reaches $\approx10^{-9}$ and is unchanged at twenty; RD converges within the default three. 
Following PCFM~\cite{utkarsh2025}, the curved cases apply one further full Gauss-Newton projection at $t = 0$ to remove the linearisation residual. 
These settings are fixed per benchmark a priori and are shared by SAPC and by every sampling-time baseline that uses a projector; no sampling-time hyperparameter is tuned against the reported test splits.

Each benchmark carries four held-out test splits. Per split we generate $1225$ trajectories at an evaluation batch size of $32$; on Navier-Stokes we generate 100 trajectories at batch size 16, the 3D rollout making the full 1225 prohibitive, and score them against the 1225 reference trajectories of the split.
Sampling one (seed, split) pair takes $3$-$15$ minutes on the 2D datasets and $\approx22$ minutes on Navier-Stokes. 
Every reported number is therefore the mean $\pm$ standard error over the $3 \times 4 = 12$ (seed, split) runs of a  cell.

\begin{table}[ht]
\centering
\caption{Sampling-time projector configuration. Affine families admit
an exact one-shot projection; the curved ones use damped Gauss-Newton with an early stop on $\lVert h \rVert_\infty$ and a final projection at $t = 0$.}
\label{tab:projector-config}
\begin{tabular}{cccccc}
\toprule
Benchmark & Projector & Max iters. & Tolerance & Damping $\mu$ & Final $t=0$ pass\\
\midrule
Heat           & affine (exact) & 1 & - & - & no \\
Burgers        & affine (exact) & 1 & - & - & no \\
NS  & affine (exact) & 1 & - & - & no \\
RD            & Gauss-Newton  & 3 & $10^{-6}$ & $10^{-8}$ & yes \\
Stokes IC / BC  & Gauss-Newton  & 10 & $10^{-6}$ & $10^{-8}$ & yes \\
\bottomrule
\end{tabular}
\end{table}

\subsection{Baseline Models}
\label{app:competitors}

All competitors share the same FFM backbone as our method, so that any performance gap reflects how the physical constraint is enforced, rather than differences in architecture, dataset, projector, or noise. 
D-Flow, ECI, and PCFM are \textit{zero-shot} (sampling-time only); 
CAFM and PBFM fold the constraint into training. 
For this work, we retrain all baselines following the authors' specification.

\paragraph{D-Flow.}
D-Flow~\citep{ben2024} enforces constraints on samples from a pretrained FFM by optimizing the source noise so that the resulting trajectory's endpoint minimizes a loss, without any change to the underlying network.
Because the objective is evaluated only at the endpoint, standard backpropagation through an unrolled Euler solver supplies the gradient of the loss with respect to the source. 
Following the D-Flow setup of~\citet{cheng2025, utkarsh2025}, an initial noise sample is drawn from the Gaussian-process source and integrated forward through $n=100$ Euler steps of the frozen FFM velocity network, yielding an endpoint estimate $\hat V_0$. 
The loss combines two squared residuals: a local term
$\lVert (\hat{\mathcal{V}}_0 - \mathcal{V}_0) \odot M \rVert^2$
pinning the IC/BC values through a Boolean mask $M$ over the constrained grid
points, equivalent to $\lVert \mathcal{R}_\text{loc}(\hat{\mathcal{V}}_0)\rVert^2$
in Section~\ref{app:projector}; and a weighted global term
$\lambda \lVert \mathcal{R}_\text{glob}(\hat{\mathcal{V}}_0) \rVert^2$.
We set $\lambda = 1$, mirroring the equal-weight concatenation of local and
global residuals used by the projection operator across all competitors.
The combined loss is minimized with respect to the source noise by L-BFGS (20 iterations, learning rate $0.1$). 
Once L-BFGS terminates, one final forward Euler integration with the optimized, detached noise produces the returned sample.

\paragraph{ECI.}
ECI~\citep{cheng2025} is a training-free, zero-shot sampler for FFM that, at every discretized time step, cycles through three operations: extrapolating the current state to an endpoint estimate $\hat V_0$ via the flow, correcting $\hat V_0$ by projecting it onto the constraint set, and interpolating between the corrected endpoint and the source sample to obtain the next state. 
Repeating this cycle several times per outer step enforces constraint satisfaction at the cost of extra network evaluations. 
We follow the sampling procedure of~\citet{cheng2025}, interleaving extrapolation, correction, and interpolation at every Euler step, with $n_{\text{mix}}=5$ mixing updates per outer step and no periodic noise resampling on all datasets. 
Unlike the reference implementation, our projector is not restricted to constraints admitting a closed-form oblique projection: on non-linear constraints, we substitute the non-linear projector used by the other competitors.

\paragraph{PCFM.}
PCFM~\citep{utkarsh2025} generalizes ECI by replacing
its closed-form linear projection with the Gauss-Newton projector (Section~\ref{app:projector}) to handle non-linear constraints.
We reproduce authors' implementation: $n_\text{iter}=1$ Gauss-Newton step
per Euler update (in root-finding mode, damping $\epsilon = 10^{-6}$); optional relaxed-penalty correction disabled; an OT-displacement reverse update in place of ECI's direct re-interpolation; and a final Newton-Schur projection of 3 iterations at $t=1$.
The last two elements (the OT-displacement update and the Newton-Schur pass at $t=1$) distinguish PCFM from ECI here, since the two competitors already share the same non-linear Gauss-Newton projector in our setup.

\paragraph{PBFM.}
PBFM~\citep{baldan2026} incorporates the governing equations in
the training loss, leaving inference unmodified. 
Two mechanisms distinguish it from a naive physics-penalty baseline. 
First, the FM loss $\mathcal L_{\mathrm{FM}}$ and a physical-residual loss $\mathcal L_{\mathcal R}$ are combined via conflict-free gradient updates rather than a scalar
weight (ConFIG, Conflict-Free Inverse Gradients). 
Denoting their gradients $g_{\mathrm{FM}}, g_{\mathcal R}$, the
orthogonal component $\mathcal O(g_1,g_2) = g_2 - \frac{g_1^\top g_2}{\lVert
g_1\rVert^2} g_1$, and normalisation $\mathcal U(g)=g/\lVert g\rVert$, the
update direction $g_v = \mathcal U[\mathcal U(\mathcal O(g_{\mathrm{FM}},
g_{\mathcal R})) + \mathcal U(\mathcal O(g_{\mathcal R}, g_{\mathrm{FM}}))]$
is rescaled by $g_{\mathrm{update}} = (g_{\mathrm{FM}}^\top g_v +
g_{\mathcal R}^\top g_v)\, g_v$, so that it never opposes either individual
gradient. 
Second, the residual is evaluated on the endpoint of an $n$-step Euler
unrolling of $v_\theta$ from $\mathcal V_t$ rather than on the one-step
estimate $\hat{\mathcal V}_0 = \mathcal V_t -t\, v_\theta(\mathcal V_t, t)$ (in the convention that data is at $t=0$).
The latter extrapolates linearly using the velocity at the current point and is exact only if $v_\theta$ is constant along the trajectory; since the field varies along the flow, the one-step endpoint lies off the clean-data manifold, and the non-linear $\mathcal R$ can be dominated by that extrapolation error rather than by genuine constraint violation.
Unrolling recomputes $v_\theta$ at each intermediate state, following the curvature of the learned ODE and producing an endpoint that mirrors the sampler's inference-time output, so $\mathcal R$ measures the physical fidelity of states the model actually generates. 
Our implementation follows the reference training procedure. 
For a sampled $t \sim \mathcal U[0,1]$ and interpolant $\mathcal V_t$, the endpoint is reached by unrolling the network for $n_{\mathrm{unroll}} \in [1,4]$ Euler steps, with $n_{\mathrm{unroll}}$ ramped linearly from $1$ to $4$ over four training stages
(their curriculum, reproduced against optimizer steps rather than epochs).
We evaluate $\mathcal{R}(\cdot)$ on the unrolled endpoint and multiply the
residual loss by $(1-t)$, which is PBFM's $t^p$ scaling at
$p=1$ rewritten in our time convention (data at $t=0$), and $p=1$ is the value they identify as optimal.
The FM and residual losses are backpropagated separately and their gradients combined via the ConFIG rule described above.
The logit-normal timestep weighting of the FM loss is left off, matching the reference default.  
At sampling time, PBFM performs plain Euler integration of the trained network with no projection or gradient guidance; we do not enable PBFM's optional stochastic sampler, matching the reference's deterministic default.

\paragraph{CAFM.}
CAFM~\citep{christopher2026} targets the training/sampling mismatch of projection-based FFM samplers with a training objective that regresses the projected prediction directly against the true endpoint, $\mathcal{L}_{\mathrm{CAFM}} = \lVert P\big(z_t + (1-t)\,v_\theta(z_t,t)\big) - z_1 \rVert^2$. 
The projection $P$ is realized as a differentiable projection layer (as in~\citet{utkarsh2025}) to manage non-linear constraints.
We implement CAFM following authors' specifications: sampling $t \sim \mathcal U[t_{\min}, 1]$ with $t_{\min} = 0.05$ (mirroring their $t \le 0.95$ under our time convention), forming the interpolant $V_t = (1-t)V_0 + tV_1$ and the network's raw endpoint estimate $\hat V_0 = V_t - t\, v_\theta(V_t, t)$, projecting it through the benchmark's differentiable projector to obtain $V_0^{\mathrm{proj}} = P(\hat V_0)$, and regressing $v_\theta$ toward the resulting projected velocity $v^{\mathrm{proj}} = (V_t - V_0^{\mathrm{proj}}) / \max(t, 10^{-3})$ against the ground-truth target $V_1 - V_0$ with an $\ell_2$ loss. 
As in the paper, the loss also adds a small residual regularizer $\lambda \lVert \mathcal{R}(V_0^{\mathrm{proj}}) \rVert^2$ with $\lambda = 10^{-3}$ on the projected endpoint. 
Sampling is done using the same hyperparameters as the PCFM baseline, so that any difference in downstream results isolates the effect of the training objective.

\subsection{Evaluation metrics}
\label{app:metrics}

\textbf{Pointwise moments.} To compare two distributions over continuous functions, following \citet{kerrigan2023} and all successive works, we compute the pixel-wise mean $\bar{\mathcal{V}}(\omega)$ and Bessel-corrected standard deviation $\sigma(\omega)$ of the generated and reference trajectories at each spatial location $\omega \in \Omega$, and report the mean squared errors between the two:
\begin{equation}
    \mathrm{MMSE} = \tfrac{1}{|\Omega|}\!\sum_{\omega \in \Omega}
    \big(\bar{\mathcal{V}}^{\text{gen}}(\omega) - \bar{\mathcal{V}}^{\text{gt}}(\omega)\big)^{2},
    \qquad
    \mathrm{SMSE} = \tfrac{1}{|\Omega|}\!\sum_{\omega \in \Omega}
    \big(\sigma^{\text{gen}}(\omega) - \sigma^{\text{gt}}(\omega)\big)^{2}
\end{equation}
Lower values indicate that the generated trajectories reproduce the first two marginal moments of the reference distribution more faithfully at every spatial location.

\textbf{Fr\'echet Poseidon Distance.} 
FPD is the Fr\'echet distance between Gaussian fits to the encoder-bottleneck activations of the pretrained PDE foundation model Poseidon-B \citep{herde2024}:
\begin{equation}
    \mathrm{FPD} = \lVert \mu_{\text{gen}} - \mu_{\text{gt}} \rVert_{2}^{2}
    + \mathrm{Tr}\!\left(\Sigma_{\text{gen}} + \Sigma_{\text{gt}}
    - 2\,(\Sigma_{\text{gen}}\Sigma_{\text{gt}})^{1/2}\right)
\end{equation}
Here $\mu$ and $\Sigma$ denote the mean and Bessel-corrected covariance of the $d$-dimensional feature vectors $\phi(\mathcal{V}) \in \mathbb{R}^{d}$ extracted from the Poseidon encoder bottleneck, computed over the generated and reference batches.
As in \citet{cheng2025}, we used the version of Poseidon with 157.7M model parameters, whose hidden activation size is $16 \times 768$, which we mean-pool into a $768$-dimensional vector for FPD computation.
For spatio-temporal datasets, FPD is computed per frame and averaged over time.
A lower FPD indicates a closer match in the learned feature space. 

\textbf{Per-pixel marginals.} FPD summarises quality of the generated trajectories at the level of aggregate features and can miss discrepancies in the marginal distribution at individual locations. 
We therefore complement it with the Wasserstein-2 distance $W_{2}$~\citep{baldan2026}, computed between the sorted empirical distributions at each spatial location $\omega \in \Omega$ and averaged over $\Omega$. 
Lower values indicate closer agreement between the generated and reference marginals at every location.

\textbf{Constraint error.}
Following~\citet{utkarsh2025}, we measure the constraint residual as the mean
$\ell^2$-norm across the $N$ generated samples, reported separately for the local and global components:
\begin{equation}
    \mathrm{CE}_L \;=\; \frac{1}{N}\sum_{n=1}^{N}
        \bigl\|\mathcal{R}_{\mathrm{loc}}(\mathcal{V}_n)\bigr\|_2,
    \qquad
    \mathrm{CE}_G \;=\; \frac{1}{N}\sum_{n=1}^{N}
        \bigl\|\mathcal{R}_{\mathrm{glob}}(\mathcal{V}_n)\bigr\|_2
\end{equation}
Lower values indicate stricter satisfaction of the constraints. Because the two are different physical quantities and can sit orders of magnitude apart on the same benchmark, we report $\mathrm{CE}$ by normalizing each by its value on the unconstrained FFM baseline before averaging:
\begin{equation}
    \mathrm{CE} \;=\; \frac{1}{2}\left(
        \frac{\mathrm{CE}_L}{\mathrm{CE}_L^{\mathrm{FFM}}}
        \;+\;
        \frac{\mathrm{CE}_G}{\mathrm{CE}_G^{\mathrm{FFM}}}
    \right)
\end{equation}
$\mathrm{CE}$ is therefore dimensionless, equals $1$ for the unconstrained baseline by construction, and reads as the fraction of baseline violation remaining. 

\subsection{Performance evaluation}
\label{app:performance}

Table~\ref{tab:inference_cost} reports the inference cost of each sampler on a single GH200. 
Two competitors set the extremes. 
PBFM matches the unconstrained FFM wall-clock ($\approx 1.00\times$FFM) since its constraint enters only the training loss and sampling is unchanged. 
D-Flow is two orders of magnitude slower ($97$-$140\times$FFM) because guided generation requires $2100$ NFE and backpropagation through the ODE trajectory at every optimization step. 
The projection-based samplers (PCFM, ECI, CAFM, SAPC) sit between these bounds, and their relative cost depends on where the projector is invoked along the trajectory and how its cost scales with the state dimension.

\begin{table}[!h]
\centering
\caption{Inference cost of each sampler, measured on one NVIDIA GH200 96GB GPU. 
NFE is the number of network forward evaluations per sample; IT is the wall-clock time per generated sample, $\times$FFM that time relative to the unconstrained FFM baseline, and T$_{\text{tot}}$ the wall-clock of the complete evaluation run ($n = 1225$ samples, $n = 100$ on Navier-Stokes). 
Each cell is the median over 3 seeds $\times$ 4 held-out test sets. }
\label{tab:inference_cost}
\setlength{\tabcolsep}{3.0pt}
\begin{tabular}{ccccccccc}
\toprule
 & \small Metric & SAPC & PCFM & ECI & CAFM & PBFM & D-Flow & FFM \\
\cmidrule(lr){3-9}
 & NFE & $100$ & $100$ & $500$ & $100$ & $100$ & $2100$ & $100$ \\
\midrule
\multirow{3}{*}{\rotatebox[origin=c]{90}{\small Heat}} & \small IT [ms] & $100$ & \underline{$43.2$} & $497$ & $44.0$ & $\bm{20.8}$ & $2739$ & $20.7$ \\
 & \small $\times$FFM & $4.85$ & $2.09$ & $24.02$ & $2.13$ & $1.00$ & $133$ &  \\
 & \small T$_{\text{tot}}$ [s] & $123$ & \underline{$52.9$} & $608$ & $53.9$ & $\bm{25.4}$ & $3356$ & $25.3$ \\
\midrule
\multirow{3}{*}{\rotatebox[origin=c]{90}{RD}} & \small IT [ms] & $127$ & \underline{$78.6$} & $862$ & $80.9$ & $\bm{25.5}$ & $2755$ & $26.0$ \\
 & \small $\times$FFM & $4.91$ & $3.03$ & $33.21$ & $3.12$ & $0.98$ & $106$ &  \\
 & \small T$_{\text{tot}}$ [s] & $156$ & \underline{$96.3$} & $1056$ & $99.1$ & $\bm{31.2}$ & $3375$ & $31.8$ \\
\midrule
\multirow{3}{*}{\rotatebox[origin=c]{90}{\small StokesIC}} & \small IT [ms] & $675$ & \underline{$96.4$} & $3295$ & $96.4$ & $\bm{21.6}$ & $2906$ & $20.8$ \\
 & \small $\times$FFM & $32.43$ & $4.63$ & $158$ & $4.63$ & $1.04$ & $140$ &  \\
 & \small T$_{\text{tot}}$ [s] & $827$ & \underline{$118$} & $4036$ & $118$ & $\bm{26.4}$ & $3560$ & $25.5$ \\
\midrule
\multirow{3}{*}{\rotatebox[origin=c]{90}{\small StokesBC}} & \small IT [ms] & $669$ & $93.0$ & $3259$ & \underline{$91.8$} & $\bm{21.7}$ & $2796$ & $20.9$ \\
 & \small $\times$FFM & $32.04$ & $4.45$ & $156$ & $4.40$ & $1.04$ & $134$ &  \\
 & \small T$_{\text{tot}}$ [s] & $819$ & $114$ & $3993$ & \underline{$113$} & $\bm{26.6}$ & $3425$ & $25.6$ \\
\midrule
\multirow{3}{*}{\rotatebox[origin=c]{90}{\small Burgers}} & \small IT [ms] & $102$ & $39.3$ & $503$ & \underline{$37.2$} & $\bm{21.9}$ & $2735$ & $21.9$ \\
 & \small $\times$FFM & $4.65$ & $1.80$ & $22.99$ & $1.70$ & $1.00$ & $125$ &  \\
 & \small T$_{\text{tot}}$ [s] & $125$ & $48.2$ & $616$ & \underline{$45.5$} & $\bm{26.8}$ & $3351$ & $26.8$ \\
\midrule
\multirow{3}{*}{\rotatebox[origin=c]{90}{\small NS}} & \small IT [ms] & $3913$ & $20401$ & \underline{$3664$} & $20414$ & $\bm{249}$ & $25131$ & $258$ \\
 & \small $\times$FFM & $15.18$ & $79.16$ & $14.22$ & $79.21$ & $0.97$ & $97.51$ &  \\
 & \small T$_{\text{tot}}$ [s] & $391$ & $2040$ & \underline{$366$} & $2041$ & $\bm{24.9}$ & $2513$ & $25.8$ \\
\bottomrule
\end{tabular}
\end{table}

\section{Ablation Study}
\label{app:ablation}

SAPC combines three elements: source projection $P(\mathcal{V}_1)$; a constraint-aware training loss that incorporates the projected source; and endpoint projection $P(\mathcal{V}_{0,\theta})$ at every reverse step. 
To isolate which of these drives the gain, we compare the full method against three ablations (named S1, S2, and S3) in which one or two components are removed, keeping the FFM backbone and the projector $P$ fixed. 
Table~\ref{tab:strategy_ablation} reports the results over the six datasets.

S3 applies source projection only during sampling, using a backbone trained on unconstrained noise.
As shown in Table~\ref{tab:strategy_ablation}, constraint errors  $\mathrm{CE}$,  $\mathrm{CE}$$_L$, and  $\mathrm{CE}$$_G$ are of the same order as those of the unconstrained FFM across all datasets.
Projecting the source onto $\mathcal{H}$ alone therefore does not ensure that the final sample satisfies the constraints, as the learned velocity field can transport samples away from the manifold, degrading constraints satisfaction.
The distributional accuracy of S3 varies instead across datasets.
These results highlight the importance of aligning the source distributions used during training and sampling.
SAPC achieves this alignment through a constraint-aware loss that learns transport from the anchored source. 
Source anchoring with matched training improves distributional accuracy, while endpoint projection enforces the constraints.

The two variants that project only the endpoint estimate (S2 and S1, with and without training) bring the constraint errors  $\mathrm{CE}$,  $\mathrm{CE}$$_L$, and  $\mathrm{CE}$$_G$ down to values comparable with the full SAPC method. 
The distributional metrics MMSE, SMSE, FPD, and $W_2$ nonetheless remain close to those of the unconstrained backbone: on Heat, RD, and Burgers, MMSE and FPD are two to three orders of magnitude worse than SAPC, while on NS the gap narrows to a factor of three to four. 
Enforcing the constraint pointwise on the endpoint estimate is therefore sufficient to make each generated field satisfy the PDE, but does not push the ensemble towards the target solution distribution.
The network produces admissible but distributionally biased samples.

Comparing S2 and S1 columns of Table~\ref{tab:strategy_ablation} isolates the effect of folding the endpoint projection into the training loss, at fixed sampling procedure. 
On all distributional metrics, the two variants are mostly indistinguishable across every benchmark; on the constraint errors they agree to within a factor of two on Heat, RD, and the two Stokes tasks, while on Burgers and NS the zero-shot variant is one to two orders of magnitude better on  $\mathrm{CE}$$_L$. 
Endpoint projection therefore does not benefit from training-time embedding. 
This identifies the pairing of source anchoring with a matched training loss as the mechanism driving SAPC's improvement.

For completeness, the last column of Table~\ref{tab:strategy_ablation} reports condFFM, a soft-conditioning FFM that receives the IC/BC as an additional input channel concatenated with the Gaussian source. 
This variant assesses the distributional accuracy achievable through conditioning alone, without projection.
On MMSE, SMSE, FPD, and $W_2$, condFFM is competitive with SAPC and occasionally better. 
This confirms that soft conditioning on the IC/BC captures much of the shape of
the constrained data distribution. 
The trade-off appears on the constraint errors: $\mathrm{CE}$$_L$ of condFFM is six to thirteen orders of magnitude worse than SAPC's on every benchmark, and  $\mathrm{CE}$$_G$ spans $10^{-2}$ to $10^{0}$ against SAPC's $10^{-8}$ to $10^{-2}$ across datasets. 
Soft conditioning is therefore a valid approach to distributional quality but leaves the physical constraint essentially unenforced, whereas SAPC addresses both.

%ablation 
\begin{table}[ht]
\centering
\caption{Ablation of SAPC over the constraint-aware training, the projection $P(\mathcal{V}_1)$ of the source, and the projection $P(\mathcal{V}_{0,\theta})$ of the endpoint estimate at every reverse step. FFM and condFFM are also reported. Each cell reports the mean $m$ and standard error $s$ across seeds and held-out test sets in the compact form $m(s)p$, meaning $(m \pm s) \times 10^{p}$. 
Lower values indicate better performance, with the best result in bold and the second best underlined.}
\label{tab:strategy_ablation}
\setlength{\tabcolsep}{1.4pt}
\begin{tabular}{ccccccc|c}
\toprule
 & 
 & SAPC & SAPC (S3) & SAPC (S2) & SAPC (S1) & FFM & condFFM \\
\cmidrule(lr){3-8}
 & \scriptsize training & $\checkmark$ & $\times$ & $\checkmark$ & $\times$ & $\times$ & $\times$ \\
 & \scriptsize conditioning & $\times$ & $\times$ & $\times$ & $\times$ & $\times$ & $\checkmark$ \\
 & \scriptsize $P(\mathcal{V}_1)$ & $\checkmark$ & $\checkmark$ & $\times$ & $\times$ & $\times$ & $\times$ \\
 & \scriptsize $P(\mathcal{V}_{0,\theta})$ & $\checkmark$ & $\times$ & $\checkmark$ & $\checkmark$ & $\times$ & $\times$ \\
\midrule
\multirow{7}{*}{\rotatebox[origin=c]{90}{\small Heat}} & \small MMSE & $\bm{2.2{\scriptstyle (0.8){-}05}}$ & \underline{$1.5{\scriptstyle (0.5){-}02}$} & $3.9{\scriptstyle (0.9){-}02}$ & $4.0{\scriptstyle (1.0){-}02}$ & $5.7{\scriptstyle (1.3){-}02}$ & $2.1{\scriptstyle (0.8){-}04}$ \\
 & \small SMSE & $\bm{2.3{\scriptstyle (0.4){-}05}}$ & \underline{$2.1{\scriptstyle (0.0){-}02}$} & $2.3{\scriptstyle (0.0){-}02}$ & $2.5{\scriptstyle (0.0){-}02}$ & $3.7{\scriptstyle (0.1){-}02}$ & $1.4{\scriptstyle (0.5){-}05}$ \\
 & \small FPD & $\bm{4.6{\scriptstyle (1.3){-}03}}$ & \underline{$1.6{\scriptstyle (0.1){+}00}$} & $1.9{\scriptstyle (0.2){+}00}$ & $1.9{\scriptstyle (0.2){+}00}$ & $2.7{\scriptstyle (0.3){+}00}$ & $2.3{\scriptstyle (0.8){-}02}$ \\
 & W$_2$ & $\bm{1.4{\scriptstyle (0.1){-}02}}$ & \underline{$1.4{\scriptstyle (0.1){-}01}$} & $1.9{\scriptstyle (0.1){-}01}$ & $1.9{\scriptstyle (0.1){-}01}$ & $2.3{\scriptstyle (0.1){-}01}$ & $1.8{\scriptstyle (0.2){-}02}$ \\
 & \small  $\mathrm{CE}$ & \underline{$2.0{\scriptstyle (0.0){-}06}$} & $6.8{\scriptstyle (0.6){-}01}$ & $\bm{1.9{\scriptstyle (0.0){-}06}}$ & $2.0{\scriptstyle (0.1){-}06}$ & $1.0{\scriptstyle (0.0){+}00}$ & $3.8{\scriptstyle (0.6){-}01}$ \\
 & \small  $\mathrm{CE}$$_L$ & $\bm{6.0{\scriptstyle (2.3){-}11}}$ & $3.9{\scriptstyle (0.2){+}00}$ & \underline{$2.8{\scriptstyle (0.4){-}08}$} & $4.0{\scriptstyle (0.5){-}08}$ & $6.6{\scriptstyle (0.5){+}00}$ & $4.7{\scriptstyle (0.1){-}02}$ \\
 & \small  $\mathrm{CE}$$_G$ & \underline{$5.9{\scriptstyle (0.0){-}06}$} & $1.1{\scriptstyle (0.2){+}00}$ & $\bm{5.7{\scriptstyle (0.1){-}06}}$ & $6.1{\scriptstyle (0.2){-}06}$ & $1.5{\scriptstyle (0.1){+}00}$ & $1.1{\scriptstyle (0.2){+}00}$ \\
\midrule
\multirow{7}{*}{\rotatebox[origin=c]{90}{RD}} & \small MMSE & $\bm{5.8{\scriptstyle (0.9){-}05}}$ & $7.6{\scriptstyle (1.1){-}02}$ & $3.3{\scriptstyle (1.1){-}02}$ & \underline{$3.3{\scriptstyle (1.1){-}02}$} & $4.2{\scriptstyle (1.2){-}02}$ & $1.1{\scriptstyle (0.2){-}04}$ \\
 & \small SMSE & $\bm{1.9{\scriptstyle (0.2){-}05}}$ & $3.1{\scriptstyle (0.1){-}02}$ & $2.6{\scriptstyle (0.0){-}02}$ & \underline{$2.6{\scriptstyle (0.0){-}02}$} & $3.2{\scriptstyle (0.1){-}02}$ & $6.5{\scriptstyle (0.8){-}05}$ \\
 & \small FPD & $\bm{5.9{\scriptstyle (2.0){-}01}}$ & $2.8{\scriptstyle (0.5){+}02}$ & $1.1{\scriptstyle (0.3){+}02}$ & \underline{$1.1{\scriptstyle (0.4){+}02}$} & $1.2{\scriptstyle (0.3){+}02}$ & $8.4{\scriptstyle (3.0){-}01}$ \\
 & W$_2$ & $\bm{7.0{\scriptstyle (0.2){-}03}}$ & $3.0{\scriptstyle (0.1){-}01}$ & $2.2{\scriptstyle (0.2){-}01}$ & \underline{$2.2{\scriptstyle (0.2){-}01}$} & $2.5{\scriptstyle (0.2){-}01}$ & $1.2{\scriptstyle (0.1){-}02}$ \\
 & \small  $\mathrm{CE}$ & $\bm{3.3{\scriptstyle (0.2){-}06}}$ & $1.7{\scriptstyle (0.1){+}00}$ & $4.3{\scriptstyle (0.1){-}06}$ & \underline{$4.2{\scriptstyle (0.1){-}06}$} & $1.0{\scriptstyle (0.0){+}00}$ & $8.2{\scriptstyle (1.2){-}01}$ \\
 & \small  $\mathrm{CE}$$_L$ & $\bm{1.9{\scriptstyle (0.2){-}14}}$ & $5.1{\scriptstyle (0.0){+}00}$ & $1.5{\scriptstyle (0.1){-}07}$ & \underline{$1.5{\scriptstyle (0.1){-}07}$} & $4.9{\scriptstyle (0.1){+}00}$ & $1.3{\scriptstyle (0.1){-}01}$ \\
 & \small  $\mathrm{CE}$$_G$ & $\bm{3.1{\scriptstyle (0.2){-}07}}$ & $1.1{\scriptstyle (0.1){-}01}$ & $4.1{\scriptstyle (0.1){-}07}$ & \underline{$3.9{\scriptstyle (0.1){-}07}$} & $4.7{\scriptstyle (0.0){-}02}$ & $7.5{\scriptstyle (1.1){-}02}$ \\
\midrule
\multirow{7}{*}{\rotatebox[origin=c]{90}{\small Stokes IC}} & \small MMSE & $\bm{3.0{\scriptstyle (1.2){-}03}}$ & $1.4{\scriptstyle (0.2){-}02}$ & \underline{$9.9{\scriptstyle (3.6){-}03}$} & $1.0{\scriptstyle (0.4){-}02}$ & $1.5{\scriptstyle (0.4){-}02}$ & $8.0{\scriptstyle (2.9){-}04}$ \\
 & \small SMSE & $\bm{1.1{\scriptstyle (0.4){-}03}}$ & $1.7{\scriptstyle (0.1){-}02}$ & \underline{$7.7{\scriptstyle (1.0){-}03}$} & $8.0{\scriptstyle (1.1){-}03}$ & $1.9{\scriptstyle (0.2){-}02}$ & $3.6{\scriptstyle (1.0){-}04}$ \\
 & \small FPD & $\bm{2.8{\scriptstyle (1.2){-}01}}$ & $5.6{\scriptstyle (1.6){+}00}$ & \underline{$2.7{\scriptstyle (1.2){+}00}$} & $2.9{\scriptstyle (1.3){+}00}$ & $7.9{\scriptstyle (2.5){+}00}$ & $1.3{\scriptstyle (0.5){-}01}$ \\
 & W$_2$ & $\bm{3.7{\scriptstyle (1.1){-}02}}$ & $1.4{\scriptstyle (0.1){-}01}$ & \underline{$1.0{\scriptstyle (0.1){-}01}$} & $1.0{\scriptstyle (0.1){-}01}$ & $1.5{\scriptstyle (0.1){-}01}$ & $1.9{\scriptstyle (0.4){-}02}$ \\
 & \small  $\mathrm{CE}$ & $\bm{6.1{\scriptstyle (3.3){-}02}}$ & $1.0{\scriptstyle (0.0){+}00}$ & $5.1{\scriptstyle (1.9){-}01}$ & \underline{$4.8{\scriptstyle (1.8){-}01}$} & $1.0{\scriptstyle (0.1){+}00}$ & $5.0{\scriptstyle (1.3){-}01}$ \\
 & \small  $\mathrm{CE}$$_L$ & $\bm{3.0{\scriptstyle (1.7){-}08}}$ & $1.9{\scriptstyle (0.1){+}00}$ & $2.8{\scriptstyle (1.0){-}07}$ & \underline{$2.4{\scriptstyle (0.9){-}07}$} & $2.2{\scriptstyle (0.2){+}00}$ & $9.1{\scriptstyle (0.5){-}02}$ \\
 & \small  $\mathrm{CE}$$_G$ & $\bm{4.6{\scriptstyle (2.5){-}02}}$ & $4.3{\scriptstyle (0.2){-}01}$ & $3.8{\scriptstyle (1.4){-}01}$ & \underline{$3.7{\scriptstyle (1.4){-}01}$} & $3.8{\scriptstyle (0.1){-}01}$ & $3.6{\scriptstyle (1.0){-}01}$ \\
\midrule
\multirow{7}{*}{\rotatebox[origin=c]{90}{\small Stokes BC}} & \small MMSE & $\bm{1.4{\scriptstyle (0.1){-}03}}$ & \underline{$6.6{\scriptstyle (2.4){-}03}$} & $1.9{\scriptstyle (0.2){-}02}$ & $1.8{\scriptstyle (0.2){-}02}$ & $5.3{\scriptstyle (0.8){-}02}$ & $3.5{\scriptstyle (1.0){-}04}$ \\
 & \small SMSE & $\bm{3.8{\scriptstyle (0.4){-}03}}$ & \underline{$4.0{\scriptstyle (0.4){-}03}$} & $1.0{\scriptstyle (0.1){-}02}$ & $1.1{\scriptstyle (0.1){-}02}$ & $3.2{\scriptstyle (0.0){-}02}$ & $4.8{\scriptstyle (2.4){-}04}$ \\
 & \small FPD & \underline{$1.7{\scriptstyle (0.2){+}00}$} & $\bm{6.0{\scriptstyle (0.9){-}01}}$ & $5.6{\scriptstyle (1.5){+}00}$ & $5.5{\scriptstyle (1.5){+}00}$ & $2.1{\scriptstyle (0.5){+}00}$ & $1.8{\scriptstyle (0.9){-}01}$ \\
 & W$_2$ & $\bm{6.1{\scriptstyle (0.3){-}02}}$ & \underline{$7.3{\scriptstyle (0.8){-}02}$} & $1.2{\scriptstyle (0.1){-}01}$ & $1.2{\scriptstyle (0.1){-}01}$ & $1.4{\scriptstyle (0.1){-}01}$ & $2.7{\scriptstyle (0.4){-}02}$ \\
 & \small  $\mathrm{CE}$ & $\bm{3.6{\scriptstyle (0.4){-}02}}$ & $1.6{\scriptstyle (0.1){+}00}$ & $1.5{\scriptstyle (0.2){-}01}$ & \underline{$1.4{\scriptstyle (0.2){-}01}$} & $1.0{\scriptstyle (0.0){+}00}$ & $4.6{\scriptstyle (0.4){-}01}$ \\
 & \small  $\mathrm{CE}$$_L$ & $\bm{1.3{\scriptstyle (0.3){-}09}}$ & $3.0{\scriptstyle (0.2){+}00}$ & $2.2{\scriptstyle (0.1){-}07}$ & \underline{$1.7{\scriptstyle (0.1){-}07}$} & $1.2{\scriptstyle (0.1){+}01}$ & $1.0{\scriptstyle (0.0){-}01}$ \\
 & \small  $\mathrm{CE}$$_G$ & $\bm{2.8{\scriptstyle (0.3){-}02}}$ & $1.2{\scriptstyle (0.0){+}00}$ & $1.1{\scriptstyle (0.1){-}01}$ & \underline{$1.0{\scriptstyle (0.1){-}01}$} & $3.8{\scriptstyle (0.1){-}01}$ & $3.4{\scriptstyle (0.3){-}01}$ \\
\midrule
\multirow{7}{*}{\rotatebox[origin=c]{90}{\small Burgers}} & \small MMSE & $\bm{6.4{\scriptstyle (1.8){-}05}}$ & \underline{$3.5{\scriptstyle (0.5){-}02}$} & $6.2{\scriptstyle (1.0){-}02}$ & $6.3{\scriptstyle (1.0){-}02}$ & $9.0{\scriptstyle (1.2){-}02}$ & $4.1{\scriptstyle (1.1){-}05}$ \\
 & \small SMSE & $\bm{5.5{\scriptstyle (1.1){-}05}}$ & $4.5{\scriptstyle (0.1){-}02}$ & \underline{$3.3{\scriptstyle (0.1){-}02}$} & $3.4{\scriptstyle (0.1){-}02}$ & $5.0{\scriptstyle (0.1){-}02}$ & $3.8{\scriptstyle (1.0){-}05}$ \\
 & \small FPD & $\bm{1.5{\scriptstyle (0.4){-}02}}$ & \underline{$2.1{\scriptstyle (0.3){+}00}$} & $2.5{\scriptstyle (0.5){+}00}$ & $2.5{\scriptstyle (0.5){+}00}$ & $2.8{\scriptstyle (0.4){+}00}$ & $7.3{\scriptstyle (1.8){-}03}$ \\
 & W$_2$ & $\bm{9.9{\scriptstyle (0.6){-}03}}$ & \underline{$2.5{\scriptstyle (0.1){-}01}$} & $2.8{\scriptstyle (0.1){-}01}$ & $2.8{\scriptstyle (0.1){-}01}$ & $3.2{\scriptstyle (0.1){-}01}$ & $9.1{\scriptstyle (0.5){-}03}$ \\
 & \small  $\mathrm{CE}$ & \underline{$1.7{\scriptstyle (0.2){-}06}$} & $6.4{\scriptstyle (0.1){-}01}$ & $2.7{\scriptstyle (0.7){-}05}$ & $\bm{1.4{\scriptstyle (0.1){-}06}}$ & $1.0{\scriptstyle (0.0){+}00}$ & $3.3{\scriptstyle (0.1){-}01}$ \\
 & \small  $\mathrm{CE}$$_L$ & $\bm{4.4{\scriptstyle (1.3){-}10}}$ & $5.5{\scriptstyle (0.1){+}00}$ & $1.1{\scriptstyle (0.3){-}06}$ & \underline{$2.9{\scriptstyle (0.3){-}08}$} & $7.6{\scriptstyle (0.1){+}00}$ & $4.5{\scriptstyle (0.2){-}02}$ \\
 & \small  $\mathrm{CE}$$_G$ & \underline{$5.0{\scriptstyle (0.5){-}06}$} & $8.3{\scriptstyle (0.3){-}01}$ & $8.2{\scriptstyle (2.0){-}05}$ & $\bm{4.1{\scriptstyle (0.2){-}06}}$ & $1.5{\scriptstyle (0.1){+}00}$ & $9.7{\scriptstyle (0.4){-}01}$ \\
\midrule
\multirow{7}{*}{\rotatebox[origin=c]{90}{\small Navier-Stokes}} & \small MMSE & $\bm{5.3{\scriptstyle (0.6){-}02}}$ & \underline{$1.4{\scriptstyle (0.1){-}01}$} & $1.8{\scriptstyle (0.1){-}01}$ & $1.9{\scriptstyle (0.1){-}01}$ & $1.8{\scriptstyle (0.1){-}01}$ & $2.5{\scriptstyle (0.4){-}02}$ \\
 & \small SMSE & $\bm{3.1{\scriptstyle (0.4){-}02}}$ & \underline{$6.0{\scriptstyle (0.1){-}02}$} & $6.9{\scriptstyle (0.3){-}02}$ & $7.0{\scriptstyle (0.3){-}02}$ & $6.9{\scriptstyle (0.3){-}02}$ & $1.4{\scriptstyle (0.2){-}02}$ \\
 & \small FPD & $\bm{1.1{\scriptstyle (0.1){+}00}}$ & \underline{$2.4{\scriptstyle (0.1){+}00}$} & $2.6{\scriptstyle (0.1){+}00}$ & $2.7{\scriptstyle (0.1){+}00}$ & $2.7{\scriptstyle (0.1){+}00}$ & $9.3{\scriptstyle (1.6){-}01}$ \\
 & W$_2$ & $\bm{2.5{\scriptstyle (0.1){-}01}}$ & \underline{$3.8{\scriptstyle (0.1){-}01}$} & $4.2{\scriptstyle (0.1){-}01}$ & $4.2{\scriptstyle (0.1){-}01}$ & $4.2{\scriptstyle (0.2){-}01}$ & $1.8{\scriptstyle (0.1){-}01}$ \\
 & \small  $\mathrm{CE}$ & \underline{$7.3{\scriptstyle (0.1){-}07}$} & $7.4{\scriptstyle (0.4){-}01}$ & $3.6{\scriptstyle (0.1){-}06}$ & $\bm{7.1{\scriptstyle (0.1){-}07}}$ & $1.0{\scriptstyle (0.0){+}00}$ & $4.4{\scriptstyle (0.4){-}01}$ \\
 & \small  $\mathrm{CE}$$_L$ & $\bm{6.3{\scriptstyle (0.9){-}09}}$ & $1.6{\scriptstyle (0.0){+}01}$ & $4.5{\scriptstyle (0.1){-}06}$ & \underline{$1.9{\scriptstyle (0.2){-}08}$} & $2.2{\scriptstyle (0.1){+}01}$ & $2.7{\scriptstyle (0.1){+}00}$ \\
 & \small  $\mathrm{CE}$$_G$ & \underline{$8.8{\scriptstyle (0.1){-}08}$} & $4.5{\scriptstyle (0.4){-}02}$ & $4.2{\scriptstyle (0.1){-}07}$ & $\bm{8.5{\scriptstyle (0.1){-}08}}$ & $6.0{\scriptstyle (0.5){-}02}$ & $4.5{\scriptstyle (0.4){-}02}$ \\
\bottomrule
\end{tabular}
\end{table}

\section{Additional Results}
\label{app:additional_results}

Table~\ref{tab:generative_performance_supp} complements the main results by reporting the Wasserstein-2 distance $W_2$ and the constraint errors $\mathrm{CE}$$_L$ and $\mathrm{CE}$$_G$. 
Three observations reinforce the previous results.
First, $W_2$ confirms the ranking obtained from MMSE, SMSE, and FPD in Table~\ref{tab:generative_performance}.
Indeed, SAPC attains the best distributional match across all datasets, one to two orders of magnitude below the closest competitor on Heat, RD, and Burgers, and by a factor of 1.6 to 3 on the two Stokes tasks and NS. 
Second, the decomposition into $\mathrm{CE}_L$ and $\mathrm{CE}_G$ distinguishes local constraint satisfaction from global conservation accuracy. 
PCFM and CAFM reduce local errors to approximately $10^{-12}$–$10^{-14}$ on Heat, RD, and Burgers.
SAPC also achieves very small local errors, although not uniformly as low, while matching or improving global constraint accuracy: it attains the lowest $\mathrm{CE}_G$ on Heat and RD and remains within a factor of two of the best result on NS and Burgers.
Stokes BC is the exception: CAFM achieves $\mathrm{CE}_G = 8.4 \times 10^{-7}$, compared with $2.8 \times 10^{-2}$ for SAPC. 
These values should be interpreted relative to the numerical NLEC residual on exact held-out solutions, which ranges from $8.5 \times 10^{-2}$ to $1.1 \times 10^{-1}$. 
SAPC falls below this reference range by a factor of approximately three to four, whereas CAFM falls about five orders of magnitude below it. Nevertheless, CAFM yields higher MMSE, FPD, and $W_2$ than SAPC, demonstrating that a smaller numerical constraint residual does not necessarily imply closer agreement with the target distribution.
Finally, the soft-constraint competitors (PBFM, D-Flow) fail to reduce either constraint errors below the unconstrained FFM level across most datasets.
This confirms that penalty-based enforcement does not translate into satisfaction of the physical law at inference time.

\begin{table}[!h]
\centering
\caption{Supplement to Table \ref{tab:generative_performance}: the Wasserstein-2 distance ($W_{2}$) and the two constraint-errors  $\mathrm{CE}$$_L$ and  $\mathrm{CE}$$_G$ (see Section~\ref{app:metrics} for their definition). Each cell reports the mean $m$ and standard error $s$ across seeds and held-out test sets in the compact form $m(s)p$, meaning $(m \pm s) \times 10^{p}$. Lower values indicate better performance, with the best result in bold and the second best underlined}
\label{tab:generative_performance_supp}
\setlength{\tabcolsep}{1.4pt}
\begin{tabular}{ccccccccc}
\toprule
 & \small Metric & SAPC & PCFM & ECI & CAFM & PBFM & D-Flow & FFM \\
\midrule
\multirow{3}{*}{\rotatebox[origin=c]{90}{\small Heat}} & W$_2$ & $\bm{1.4{\scriptstyle (0.1){-}02}}$ & $2.6{\scriptstyle (0.1){-}01}$ & $2.6{\scriptstyle (0.5){-}01}$ & $3.3{\scriptstyle (0.5){-}01}$ & $2.3{\scriptstyle (0.2){-}01}$ & $2.3{\scriptstyle (1.0){+}00}$ & \underline{$2.3{\scriptstyle (0.1){-}01}$} \\
 & \small  $\mathrm{CE}$$_L$ & $6.0{\scriptstyle (2.3){-}11}$ & $\bm{7.7{\scriptstyle (4.2){-}14}}$ & $6.2{\scriptstyle (0.1){-}08}$ & \underline{$3.9{\scriptstyle (2.1){-}13}$} & $6.6{\scriptstyle (0.5){+}00}$ & $3.1{\scriptstyle (0.8){+}00}$ & $6.6{\scriptstyle (0.5){+}00}$ \\
 & \small  $\mathrm{CE}$$_G$ & $\bm{5.9{\scriptstyle (0.0){-}06}}$ & \underline{$6.0{\scriptstyle (0.1){-}06}$} & $7.9{\scriptstyle (0.4){-}06}$ & $6.6{\scriptstyle (0.4){-}06}$ & $4.3{\scriptstyle (0.0){-}01}$ & $5.8{\scriptstyle (2.7){+}00}$ & $1.5{\scriptstyle (0.1){+}00}$ \\
\midrule
\multirow{3}{*}{\rotatebox[origin=c]{90}{RD}} & W$_2$ & $\bm{7.0{\scriptstyle (0.2){-}03}}$ & $3.2{\scriptstyle (0.2){-}01}$ & \underline{$1.9{\scriptstyle (0.1){-}01}$} & $2.7{\scriptstyle (0.1){-}01}$ & $2.5{\scriptstyle (0.2){-}01}$ & $3.2{\scriptstyle (1.8){+}00}$ & $2.5{\scriptstyle (0.2){-}01}$ \\
 & \small  $\mathrm{CE}$$_L$ & $\bm{1.9{\scriptstyle (0.2){-}14}}$ & $1.1{\scriptstyle (0.1){-}12}$ & $1.4{\scriptstyle (0.0){-}07}$ & \underline{$2.2{\scriptstyle (0.2){-}13}$} & $4.9{\scriptstyle (0.1){+}00}$ & $5.6{\scriptstyle (2.0){+}00}$ & $4.9{\scriptstyle (0.1){+}00}$ \\
 & \small  $\mathrm{CE}$$_G$ & $\bm{3.1{\scriptstyle (0.2){-}07}}$ & $3.1{\scriptstyle (0.2){-}07}$ & $3.4{\scriptstyle (0.1){-}07}$ & \underline{$3.1{\scriptstyle (0.1){-}07}$} & $1.7{\scriptstyle (0.2){-}02}$ & $2.8{\scriptstyle (2.5){+}00}$ & $4.7{\scriptstyle (0.0){-}02}$ \\
\midrule
\multirow{3}{*}{\rotatebox[origin=c]{90}{\small StokesIC}} & W$_2$ & $\bm{3.7{\scriptstyle (1.1){-}02}}$ & $1.7{\scriptstyle (0.0){-}01}$ & \underline{$1.1{\scriptstyle (0.2){-}01}$} & $1.8{\scriptstyle (0.0){-}01}$ & $1.5{\scriptstyle (0.1){-}01}$ & $4.9{\scriptstyle (2.4){-}01}$ & $1.5{\scriptstyle (0.1){-}01}$ \\
 & \small  $\mathrm{CE}$$_L$ & $\bm{3.0{\scriptstyle (1.7){-}08}}$ & $1.0{\scriptstyle (0.2){-}06}$ & $1.3{\scriptstyle (0.2){-}07}$ & \underline{$7.1{\scriptstyle (1.7){-}08}$} & $2.2{\scriptstyle (0.2){+}00}$ & $8.2{\scriptstyle (2.0){-}01}$ & $2.2{\scriptstyle (0.2){+}00}$ \\
 & \small  $\mathrm{CE}$$_G$ & \underline{$4.6{\scriptstyle (2.5){-}02}$} & $4.7{\scriptstyle (0.7){-}01}$ & $1.4{\scriptstyle (0.4){-}01}$ & $\bm{3.8{\scriptstyle (1.2){-}02}}$ & $1.5{\scriptstyle (0.1){-}01}$ & $7.8{\scriptstyle (6.4){+}00}$ & $3.8{\scriptstyle (0.1){-}01}$ \\
\midrule
\multirow{3}{*}{\rotatebox[origin=c]{90}{\small StokesBC}} & W$_2$ & $\bm{6.1{\scriptstyle (0.3){-}02}}$ & $5.6{\scriptstyle (2.2){-}01}$ & \underline{$1.0{\scriptstyle (0.1){-}01}$} & $1.7{\scriptstyle (0.0){-}01}$ & $1.5{\scriptstyle (0.1){-}01}$ & $2.3{\scriptstyle (0.9){+}00}$ & $1.4{\scriptstyle (0.1){-}01}$ \\
 & \small  $\mathrm{CE}$$_L$ & \underline{$1.3{\scriptstyle (0.3){-}09}$} & $3.4{\scriptstyle (0.3){-}07}$ & $5.0{\scriptstyle (0.4){-}08}$ & $\bm{1.4{\scriptstyle (0.3){-}10}}$ & $1.2{\scriptstyle (0.0){+}01}$ & $5.8{\scriptstyle (1.4){+}00}$ & $1.2{\scriptstyle (0.1){+}01}$ \\
 & \small  $\mathrm{CE}$$_G$ & $2.8{\scriptstyle (0.3){-}02}$ & $3.5{\scriptstyle (0.3){-}01}$ & \underline{$2.1{\scriptstyle (0.3){-}02}$} & $\bm{8.4{\scriptstyle (1.3){-}07}}$ & $1.2{\scriptstyle (0.0){-}01}$ & $1.8{\scriptstyle (1.6){+}02}$ & $3.8{\scriptstyle (0.1){-}01}$ \\
\midrule
\multirow{3}{*}{\rotatebox[origin=c]{90}{\small Burgers}} & W$_2$ & $\bm{9.9{\scriptstyle (0.6){-}03}}$ & $4.1{\scriptstyle (0.3){-}01}$ & $3.7{\scriptstyle (0.6){-}01}$ & $2.9{\scriptstyle (0.1){-}01}$ & $3.2{\scriptstyle (0.2){-}01}$ & \underline{$1.2{\scriptstyle (0.2){-}01}$} & $3.2{\scriptstyle (0.1){-}01}$ \\
 & \small  $\mathrm{CE}$$_L$ & $4.4{\scriptstyle (1.3){-}10}$ & $\bm{0.0{\scriptstyle (0.0){+}00}}$ & $7.5{\scriptstyle (0.4){-}08}$ & \underline{$4.4{\scriptstyle (3.1){-}12}$} & $7.5{\scriptstyle (0.2){+}00}$ & $1.4{\scriptstyle (0.0){+}00}$ & $7.6{\scriptstyle (0.1){+}00}$ \\
 & \small  $\mathrm{CE}$$_G$ & $5.0{\scriptstyle (0.5){-}06}$ & \underline{$4.7{\scriptstyle (0.1){-}06}$} & $6.2{\scriptstyle (0.5){-}06}$ & $\bm{4.2{\scriptstyle (0.1){-}06}}$ & $4.2{\scriptstyle (0.2){-}01}$ & $7.6{\scriptstyle (1.1){-}01}$ & $1.5{\scriptstyle (0.1){+}00}$ \\
\midrule
\multirow{3}{*}{\rotatebox[origin=c]{90}{\small NS}} & W$_2$ & $\bm{2.5{\scriptstyle (0.1){-}01}}$ & $5.6{\scriptstyle (0.1){-}01}$ & $7.8{\scriptstyle (0.0){-}01}$ & $5.5{\scriptstyle (0.1){-}01}$ & \underline{$4.2{\scriptstyle (0.2){-}01}$} & $1.2{\scriptstyle (0.9){+}00}$ & $4.2{\scriptstyle (0.2){-}01}$ \\
 & \small  $\mathrm{CE}$$_L$ & \underline{$6.3{\scriptstyle (0.9){-}09}$} & $\bm{3.8{\scriptstyle (0.7){-}12}}$ & $6.2{\scriptstyle (0.0){-}07}$ & $4.5{\scriptstyle (1.7){-}08}$ & $2.1{\scriptstyle (0.1){+}01}$ & $1.4{\scriptstyle (0.7){+}01}$ & $2.2{\scriptstyle (0.1){+}01}$ \\
 & \small  $\mathrm{CE}$$_G$ & $8.8{\scriptstyle (0.1){-}08}$ & $\bm{8.4{\scriptstyle (0.0){-}08}}$ & $8.5{\scriptstyle (0.1){-}08}$ & \underline{$8.5{\scriptstyle (0.1){-}08}$} & $4.3{\scriptstyle (0.3){-}02}$ & $10.0{\scriptstyle (3.1){-}02}$ & $6.0{\scriptstyle (0.5){-}02}$ \\
\bottomrule
\end{tabular}
\end{table}

\begin{figure} [!h]
    \centering
    \includegraphics[width=\textwidth]{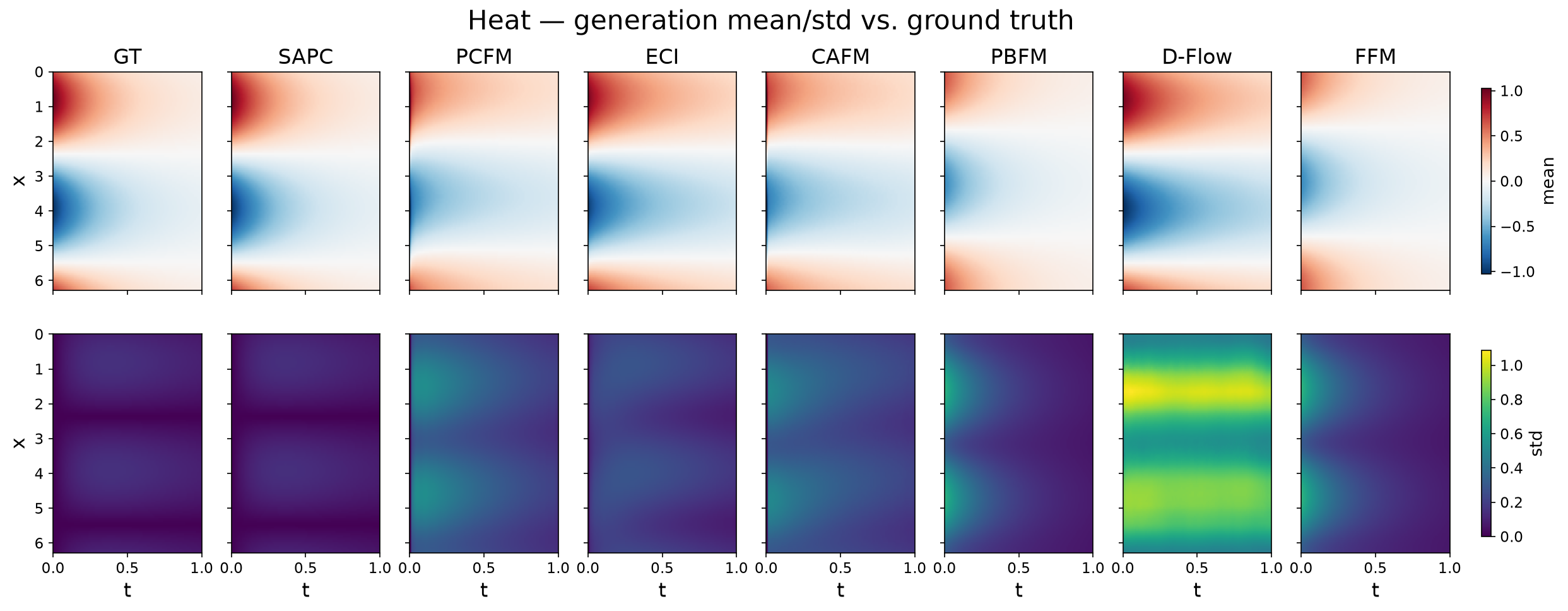}
    \caption{Generated trajectories on Heat for one test split: pointwise mean (top) and standard deviation (bottom) over $1225$ trajectories sharing the split's IC at $\phi=\pi/4$.}
    \label{fig:baseline_comparison_heat}
\end{figure}

\begin{figure} [!h]
    \centering
    \includegraphics[width=\textwidth]{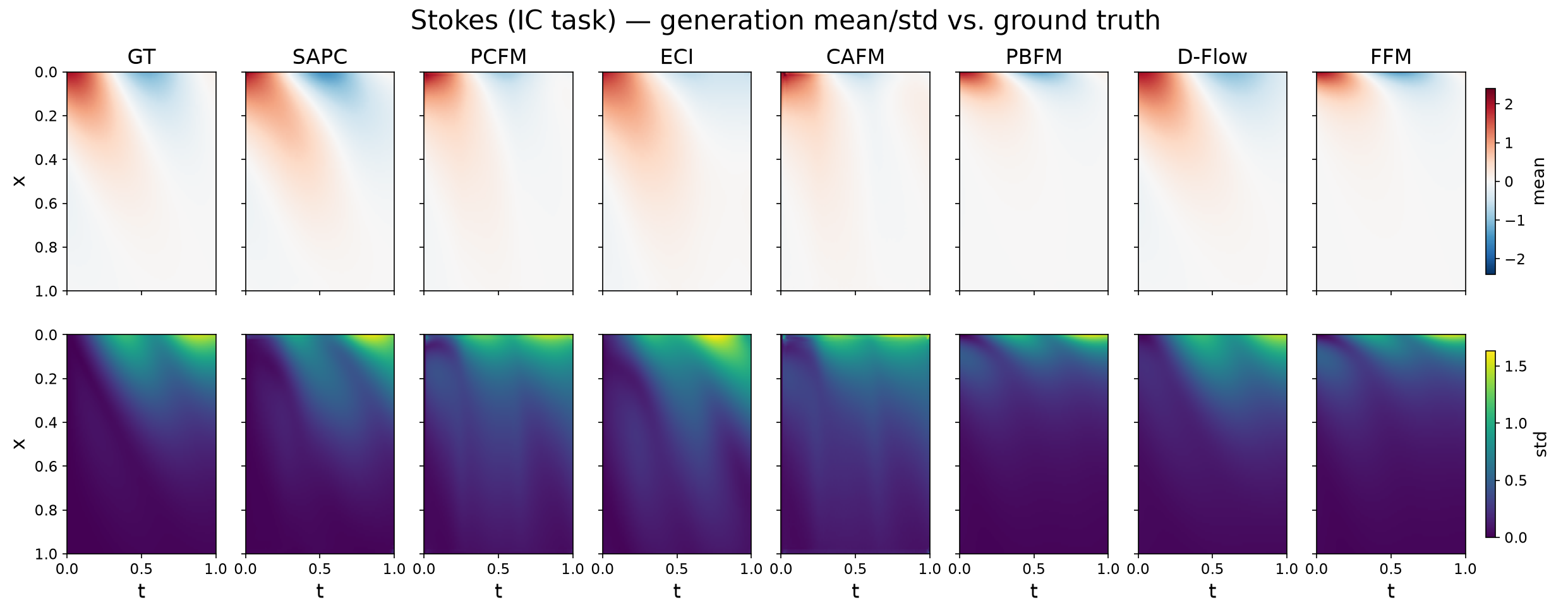}
    \caption{Generated trajectories on Stokes for one test split: pointwise mean (top) and standard deviation (bottom) over $1225$ trajectories sharing the split's IC at $k=4$, differing in forcing frequency.}
    \label{fig:baseline_comparison_stokes_ic}
\end{figure}

\begin{figure} [!h]
    \centering
    \includegraphics[width=\textwidth]{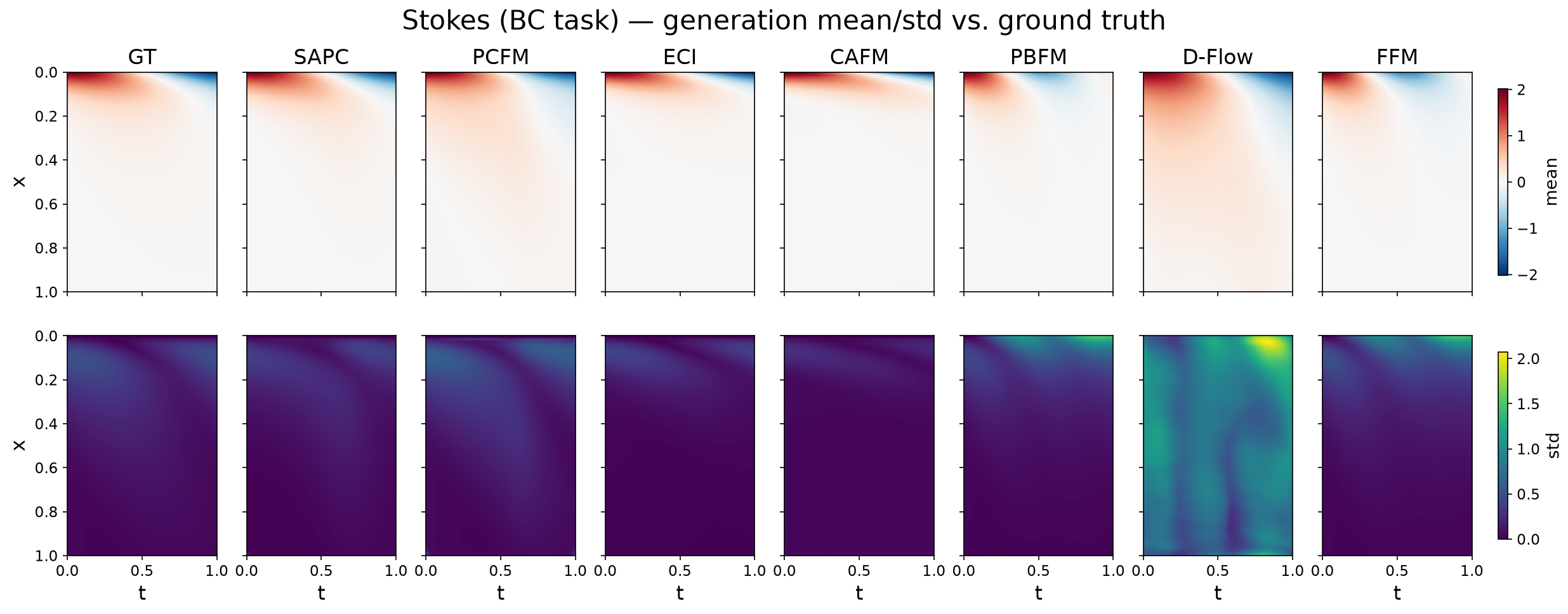}
    \caption{Generated trajectories on Stokes for one test split: pointwise mean (top) and standard deviation (bottom) over $1225$ trajectories sharing the split's BC at $\omega=3$, differing in decay rate.}
    \label{fig:baseline_comparison_stokes_bc}
\end{figure}

\begin{figure} [!h]
    \centering
    \includegraphics[width=\textwidth]{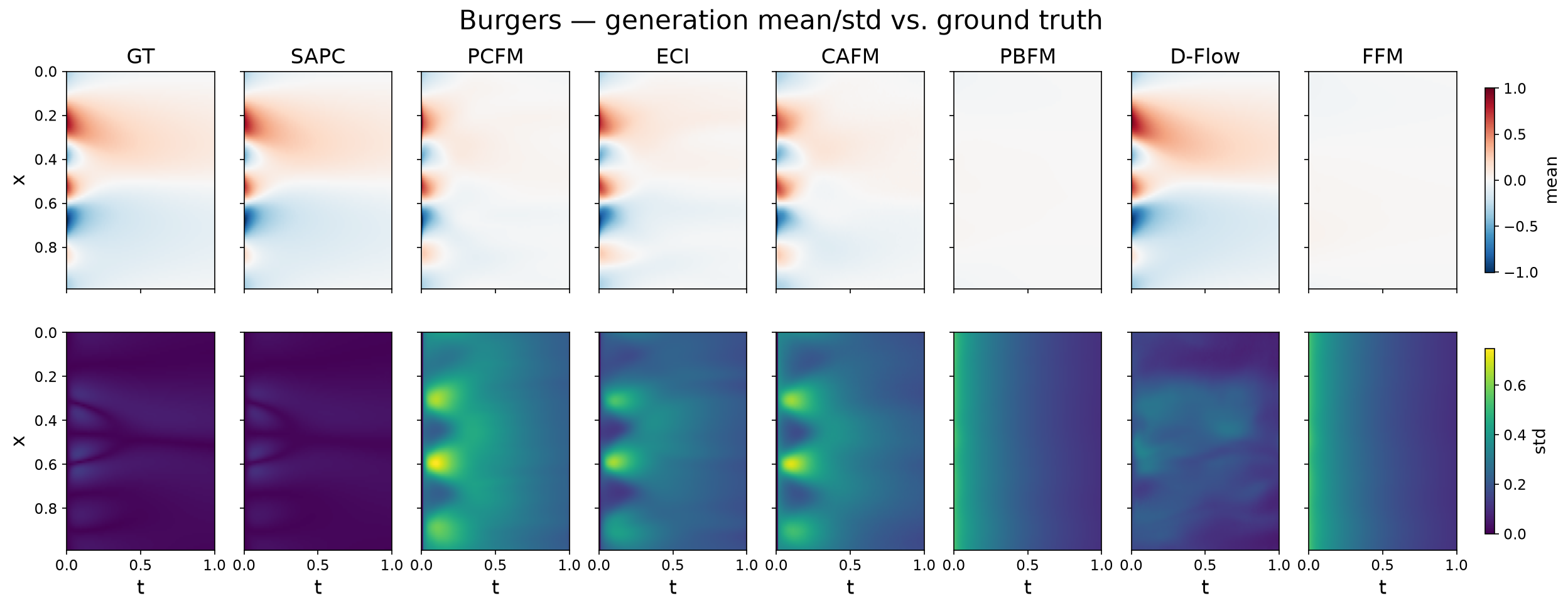}
    \caption{Generated trajectories on Burgers for one test split: pointwise mean (top) and standard deviation (bottom) over $1225$ trajectories sharing the split's IC.}
    \label{fig:baseline_comparison_burgers}
\end{figure}

\begin{figure} [!h]
    \centering
    \includegraphics[width=\textwidth]{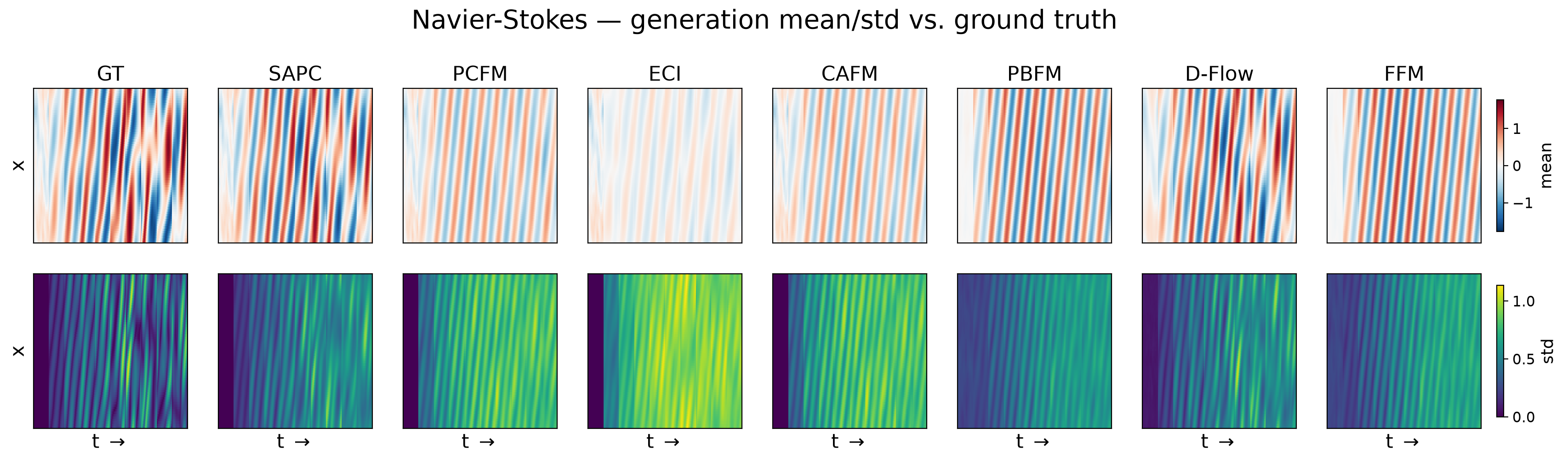}
    \caption{Generated trajectories on Navier-Stokes for one test split. Each column tiles $10$ vorticity frames left to right in $t$; trajectories share the initial vorticity and differ in the forcing phase.}
    \label{fig:baseline_comparison_ns}
\end{figure}

%competitors comparisons:
Figures~\ref{fig:baseline_comparison_heat}–\ref{fig:baseline_comparison_ns} provide the same comparison of pointwise means and standard deviations as Figure~\ref{fig:baseline_comparison_rd} for the remaining datasets. 
In each panel, the test trajectories share the prescribed IC or BC and differ only in the remaining free parameter. 
The ground-truth (GT) standard deviation therefore captures the variability under that fixed condition. SAPC and the projection-based samplers (PCFM, ECI, and CAFM) enforce the prescribed IC or BC exactly, while D-Flow approximates it through guidance. PBFM and FFM impose no constraints during sampling, allowing the IC or BC to vary according to the training distribution and consequently overestimating the pointwise standard deviation. 

\begin{figure}
    \centering
    \includegraphics[width=\textwidth]{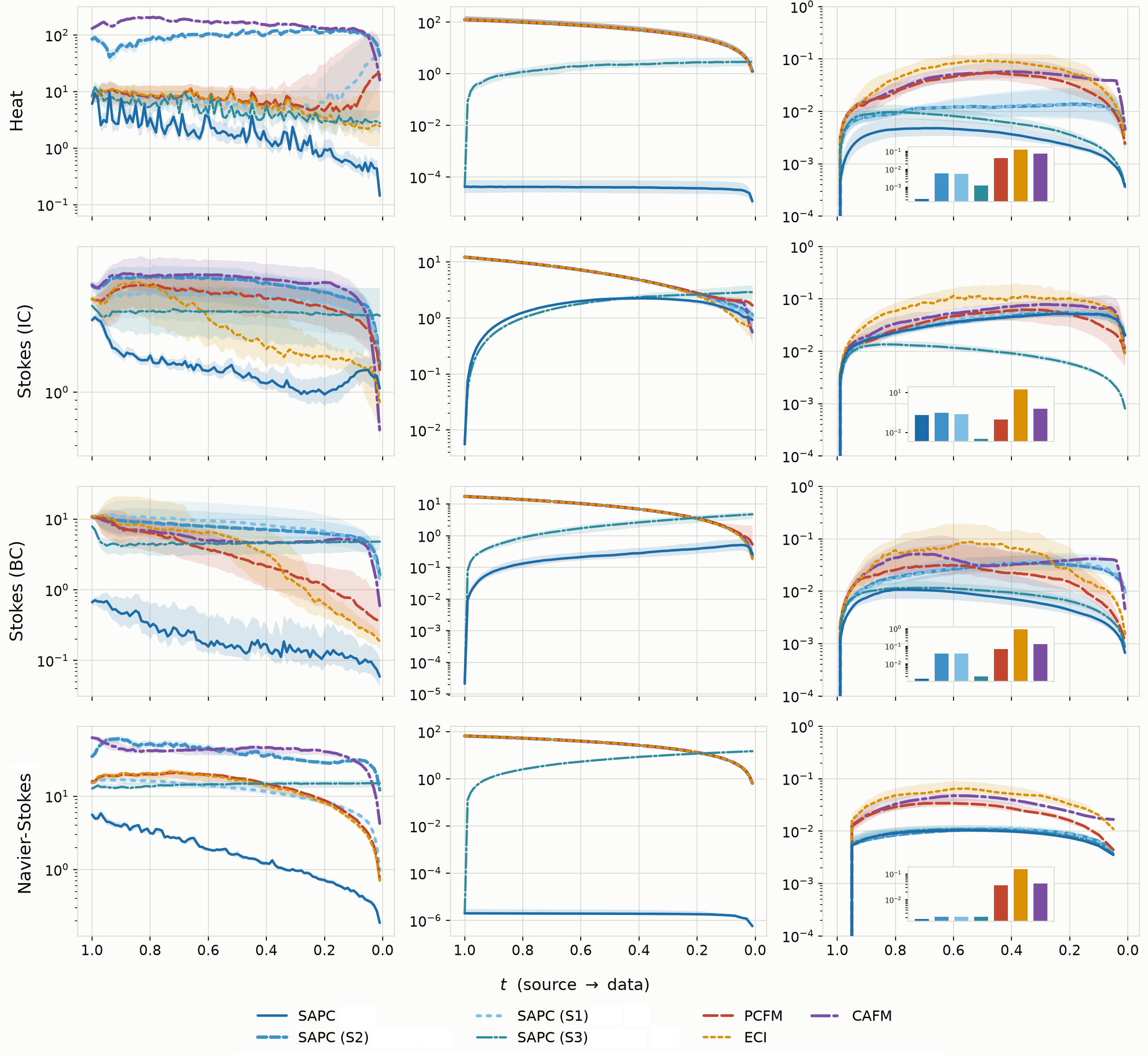}
    \caption{The same as Figure~\ref{fig:trajectory_analysis}, but for the Heat, Stokes (IC and BC) and Navier-Stokes datasets.}
    \label{fig:traj_analysis_suppl}
\end{figure}

Figure~\ref{fig:traj_analysis_suppl} extends the trajectory analysis on Heat, Stokes (IC), Stokes (BC) and NS. 
The left column shows that SAPC produces endpoint estimates $\mathcal{V}_{0,\theta}$ with the lowest constraint residual throughout sampling on Heat, Stokes~BC, and NS.
On Stokes~IC, SAPC reduces the residual by a smaller factor of two to three over most of the trajectory. Near $t=0$, however, ECI and CAFM reach lower residuals of $0.84$ and $0.60$, respectively, compared with $1.05$ for SAPC. These values describe endpoint estimates before projection, indicating that ECI and CAFM predict states that satisfy the constraints more closely at this stage. Nevertheless, projecting these estimates enforces the prescribed constraints without guaranteeing agreement with the target distribution.
The middle column is consistent with Equation~\ref{eq:sapc-nonlinear-interp}. 
On Heat and NS, where $\mathcal{R}$ is affine, SAPC maintains a low, nearly constant residual throughout sampling. 
On Stokes, the constraints combine a linear IC or BC condition with a non-linear energy balance, and the residual grows from initially small values.
Unlike RD, Stokes does not exhibit a symmetric residual profile. 
Although the Gauss–Newton projector converges to a small residual, subsequent sampling updates can move intermediate states away from the curved constraint manifold. That is the reason why, following~\citet{utkarsh2025}, we apply an additional projection at $t=0$ to remove residual constraint violations arising from non-linear projection approximations and numerical integration.
The right column shows that SAPC achieves the smallest deviation $\Delta$ and excess $\mathrm{KE}$ among the competing methods on Heat, Stokes~BC, and NS. On NS, however, SAPC and its three ablated variants (S1–S3) have similar excess $\mathrm{KE}$ ($\approx2\times10^{-3}$), suggesting that source anchoring improves constraint satisfaction without substantially changing the trajectory geometry.
Stokes~IC is the exception: SAPC achieves lower excess $\mathrm{KE}$ ($0.20$) than CAFM ($0.61$) and ECI ($17$), but higher than PCFM ($0.09$). S3 attains the lowest value ($3.2\times10^{-3}$), despite its larger constraint violations.
This is consistent with the larger iterate residual in the middle column, suggesting that maintaining constraint satisfaction on Stokes~IC requires greater deviation from the linear interpolant.

\end{document}